\pdfoutput=1
\documentclass[11pt]{article}
\usepackage{booktabs}
\usepackage[preprint]{acl}
\usepackage{datasheet}
\usepackage{times}
\usepackage{latexsym}
\usepackage{makecell}
\usepackage{graphicx}

\usepackage[T1]{fontenc}

\usepackage[utf8]{inputenc}

\usepackage{microtype}

\usepackage{inconsolata}

\usepackage{graphicx}

\title{MANBench: Is Your Multimodal Model Smarter than Human?}

\author{
 \textbf{Han Zhou\textsuperscript{1}},
 \textbf{Qitong Xu\textsuperscript{2}},
 \textbf{Yiheng Dong\textsuperscript{1}},
 \textbf{Xin Yang\textsuperscript{1}\thanks{Corresponding author.}}
\\
\\
 \textsuperscript{1}\small{School of Electronic Information and Communications, Huazhong University of Science and Technology}
 \\
 \textsuperscript{2}\small{School of Basic Medicine, Huazhong University of Science and Technology}
\\
 \small{
   {\{micdz, esoox, yiheng, xinyang2014\}@hust.edu.cn}
 }
}

\begin{document}
\maketitle

\begin{abstract}
  The rapid advancement of Multimodal Large Language Models (MLLMs) has ignited discussions regarding their potential to surpass human performance in multimodal tasks. In response, we introduce \textbf{MANBench} (Multimodal Ability Norms Benchmark), a bilingual benchmark (English and Chinese) comprising 1,314 questions across nine tasks, spanning knowledge-based and non-knowledge-based domains. MANBench emphasizes intuitive reasoning, seamless cross-modal integration, and real-world complexity, providing a rigorous evaluation framework.
  
  Through extensive human experiments involving diverse participants, we compared human performance against state-of-the-art MLLMs. The results indicate that while MLLMs excel in tasks like Knowledge and Text-Image Understanding, they struggle with deeper cross-modal reasoning tasks such as Transmorphic Understanding, Image Consistency, and Multi-image Understanding. Moreover, both humans and MLLMs face challenges in highly complex tasks like Puzzles and Spatial Imagination.
  
  MANBench highlights the strengths and limitations of MLLMs, revealing that even advanced models fall short of achieving human-level performance across many domains. We hope MANBench will inspire efforts to bridge the gap between MLLMs and human multimodal capabilities.  The code and dataset are available at \href{https://github.com/micdz/MANBench/}{https://github.com/micdz/MANBench}.
\end{abstract}
\section{Introduction}
Recent advancements in Large Language Models (LLMs) have demonstrated exceptional capabilities in text comprehension, reasoning, and generation~\citep{liu2024deepseek,dubey2024llama,jaech2024openai}. Building on this foundation, Multimodal Large Language Models (MLLMs) have emerged, significantly enhancing the ability to process and generate multimodal content~\citep{gpt4o,qwen2.5-VL,claude,team2023gemini}. These models have achieved substantial success in tasks such as image captioning, visual question answering, and visual dialog, showcasing their potential for multimodal understanding and generation.

As artificial intelligence continues to evolve rapidly, a critical question arises: Can the general multimodal capabilities of MLLMs eventually surpass those of humans? While humans possess sophisticated multimodal abilities, existing benchmarks for evaluating these capabilities exhibit several significant limitations. First, many benchmarks fail to comprehensively assess the full spectrum of multimodal reasoning, often focusing on narrow tasks or requiring domain-specific knowledge~\citep{lu2024mathvista, chen2024gmai, zhang2024mathverse, wang2024measuring}. Second, human evaluations in these benchmarks are frequently inadequate, relying on small sample sizes and limited diversity. These shortcomings hinder robust conclusions about human performance~\citep{zhang2024mathverse, wang2024measuring, yue2023mmmu, ImageNet,radford2021learning}.

To address these issues, we introduce \textbf{MANBench} (Multimodal Ability Norms Benchmark), a comprehensive benchmark designed to evaluate the multimodal capabilities of both humans and MLLMs. MANBench comprises nine tasks, encompassing 1,314 questions and 2,231 images. It aims to provide a fair and rigorous assessment framework, ensuring equitable comparisons between human and MLLM performance. Specifically, our benchmark addresses the following key limitations of existing benchmarks: (1) it separates questions requiring prior knowledge from those that do not, (2) it ensures that all questions demand reasoning rather than simple retrieval, (3) it mandates the integration of textual and visual information, and (4) it includes a wide range of reasoning difficulties to challenge both humans and MLLMs.

We conducted a large-scale human evaluation using the MANBench dataset, systematically comparing the performance of humans and MLLMs across a variety of multimodal tasks. Twelve MLLMs, including state-of-the-art models such as GPT-4o~\citep{gpt4o}, Qwen2.5-VL-72B-Instruct~\citep{qwen2.5-VL}, and Gemini-1.5 Pro~\citep{team2023gemini}, were evaluated on MANBench. Additionally, we recruited 575 participants from diverse backgrounds to complete the tasks. Our results reveal that while MLLMs excel in tasks such as Knowledge and Text-Image Understanding, they struggle with tasks requiring deeper cross-modal reasoning, including Transmorphic Understanding, Image Consistency, and Multi-image Understanding. Notably, both humans and MLLMs encounter challenges in highly complex tasks such as Puzzles and Spatial Imagination.

In summary, our contributions are as follows: (1) We introduce MANBench, a bilingual benchmark with 1,314 questions across nine tasks. (2) We conduct extensive human experiments with 575 diverse participants. (3) We evaluate 12 state-of-the-art MLLMs. (4) We provide insights into the strengths and limitations of MLLMs, highlighting the challenges in bridging the gap between MLLMs and human multimodal abilities.
\section{Related Work}
\subsection{Multimodal Large Language Models}
As LLMs advance, integrating visual content has become a critical step in the development of interactive intelligent assistants. MLLMs focus on aligning vision and language modalities. CLIP~\citep{radford2021learning} is foundational in this field, using contrastive learning on image-text pairs. Approaches like MiniGPT-4~\citep{zhu2023minigpt} and LLaVA~\citep{liu2024improved,liu2024llava}, connect vision encoders and LLMs using techniques like Q-Former and MLP, leveraging frozen
language models with a limited number of trainable parameters, demonstrating promising results. Furthermore, a range of MLLMs trained on vast datasets are now accessible to the public, pushing the boundaries of visual comprehension and generalization, including InternVL1.5, InternVL2.5 series ~\citep{chen2024internvl} and Qwen-VL~\citep{wang2024qwen2}.

\subsection{Evaluations of MLLMs}
Reasoning ability is a fundamental indicator of intelligence. Numerous benchmarks have been developed to evaluate reasoning in text-only modalities, including MathQA~\citep{amini2019mathqa}, GSM1k~\citep{zhang2024careful}, and LiveBench~\citep{livebench}. These benchmarks present questions that require advanced reasoning and knowledge, making them difficult for most people to solve.

\begin{figure*}[t]
    \centering
    \includegraphics[width=0.75\textwidth]{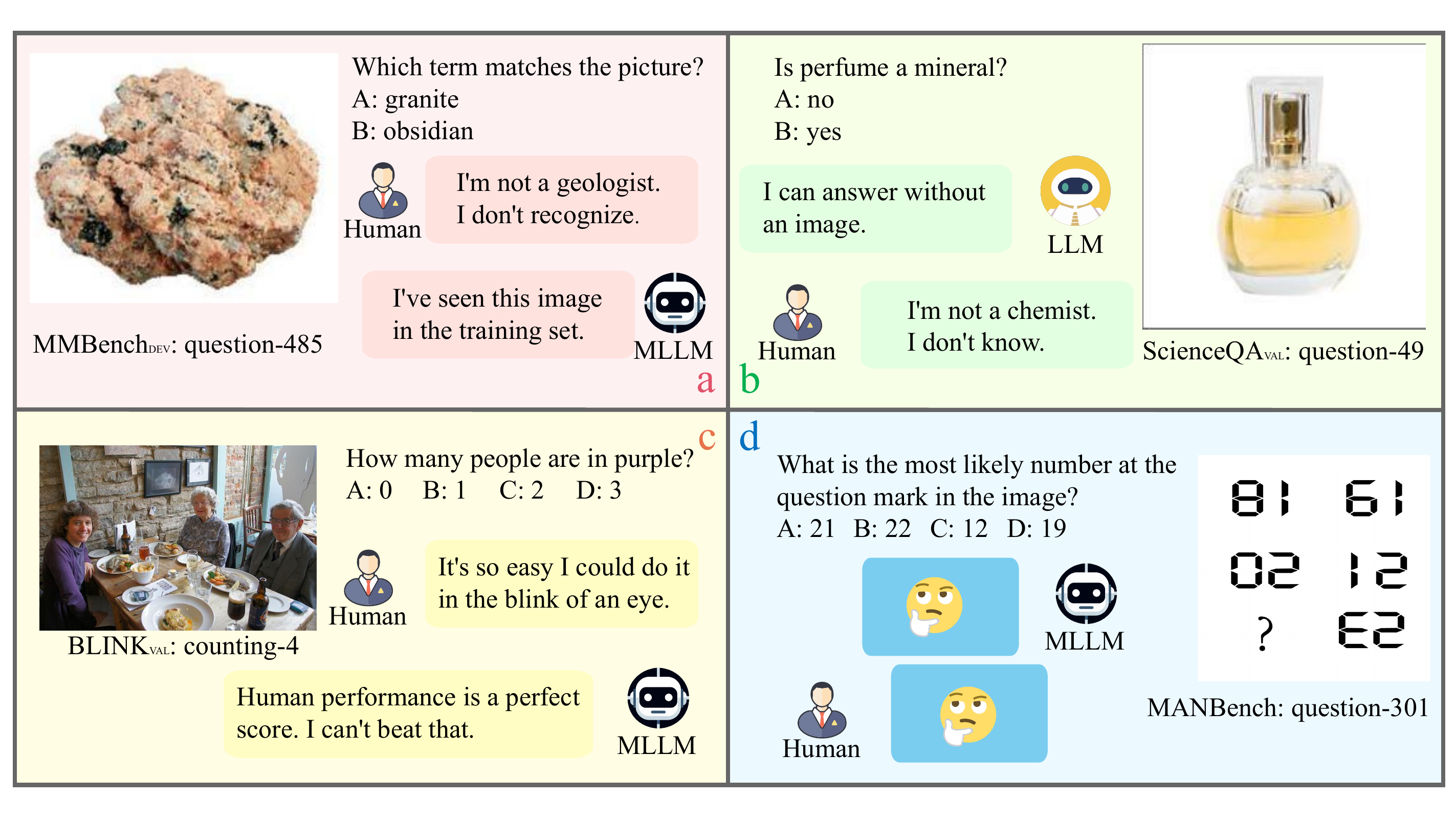}
    \caption{Limitations of existing text-image benchmarks. (a) Some samples necessitate prior knowledge rather than reasoning capabilities~\citep{MMbench}. (b)  Certain samples can be answered accurately without utilizing the image content~\citep{lu2022learn}. (c) Other samples lack sufficient reasoning complexity~\citep{fu2024blink}. (d) In contrast, our proposed questions require comprehensive reasoning, effective integration of textual and visual information, and a diverse range of reasoning difficulties.}
    
    \label{fig:related-work}
\end{figure*}

To comprehensively assess the multimodal reasoning capabilities of both humans and MLLMs, we propose the development of a novel text-image benchmark that addresses certain limitations in existing evaluation frameworks. As illustrated in Figure~\ref{fig:related-work}, current benchmarks exhibit four primary areas for improvement.

\textbf{Limitation 1: Overreliance on Prior Knowledge in Certain Samples}
Existing benchmarks frequently emphasize prior knowledge over reasoning capability \citep{lu2024mathvista, chen2024gmai, zhang2024mathverse, wang2024measuring}, leading to variability in evaluating human performance. To address this issue, MANBench introduces a clear distinction between Knowledge-based and non-knowledge-based tasks, facilitating a more balanced evaluation of cognitive processes.

\textbf{Limitation 2: Potential for Knowledge Retrieval Without Reasoning}
Current benchmarks sometimes enable solutions through direct knowledge retrieval~\citep{chen2024we}, which limits their ability to effectively evaluate reasoning skills. Our proposed dataset mitigates this with carefully constructed questions that explicitly require reasoning processes, while clearly distinguishing knowledge-based tasks.

\textbf{Limitation 3: Insufficient Integration of Multimodal Information}
Some existing benchmarks fail to effectively utilize the integration potential of textual and visual data \citep{chen2024we}. Our framework addresses this by introducing tasks that necessitate the concurrent analysis of both modalities, providing a more comprehensive assessment of multimodal reasoning capabilities.

\textbf{Limitation 4: Narrow Difficulty Spectrum}
Current benchmarks often exhibit a limited range of difficulty levels, potentially underestimating the abilities of both humans and models \citep{fu2024blink, meng2024mmiu}. To mitigate this, our dataset introduces a meticulously balanced range of reasoning complexities, enabling a more thorough evaluation across varying levels of difficulty.

\subsection{Evaluations of Humans}
Several researchers have conducted experiments to evaluate human multimodal capabilities. In the early stages of research, studies primarily focused on human image processing abilities, such as image classification~\citep{ImageNet} and few-shot classification~\citep{radford2021learning}, rather than the integrated capabilities between images and text.

Recently, there has been a surge in studies assessing human multimodal capabilities within the domain of Visual Question Answering (VQA). These studies explore specialized areas such as MathVerse~\citep{zhang2024mathverse}, MATH-Vision~\citep{wang2024measuring}, and MMMU~\citep{yue2023mmmu}. However, due to their specialized nature, it is difficult to generalize these tests to the broader population when evaluating overall human multimodal capabilities. The reliance on domain-specific knowledge implies that most questions can only be effectively answered by experts in the respective fields.

A critical limitation of both early and recent studies is the relatively small sample sizes, which may restrict the representativeness of their findings. To ensure the reliability and robustness of conclusions regarding human capabilities, studies typically require larger and more diverse participant pools to account for variability and ensure findings are broadly representative.
\section{The Multimodal Ability Norms Benchmark}

\begin{figure*}[t]
  \centering
  \includegraphics[width=0.9\textwidth]{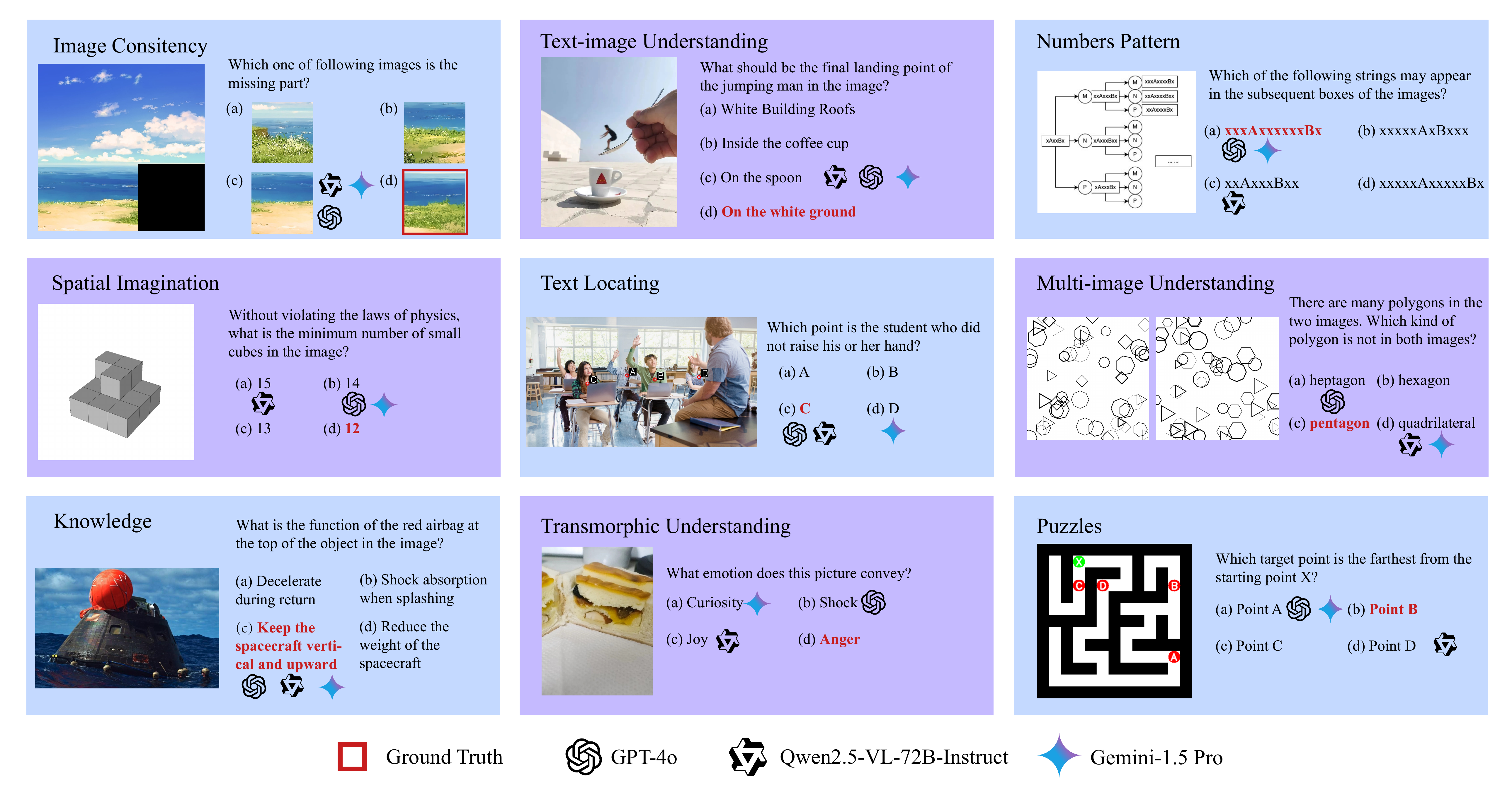}
  \caption{Qualitative results on MANBench. For each task, we display the selections made by GPT-4o, Qwen2.5-VL-72B-Instruct, and Gemini-1.5 Pro, with the red-highlighted choice indicating the ground truth.}
  \label{fig:demo} 
\end{figure*}

Our goal is to develop a benchmark that accurately evaluates the multimodal capabilities of both humans and MLLMs. 

\subsection{Overview of MANBench}
\begin{figure}[h]
    \centering
    \includegraphics[width=0.8\columnwidth]{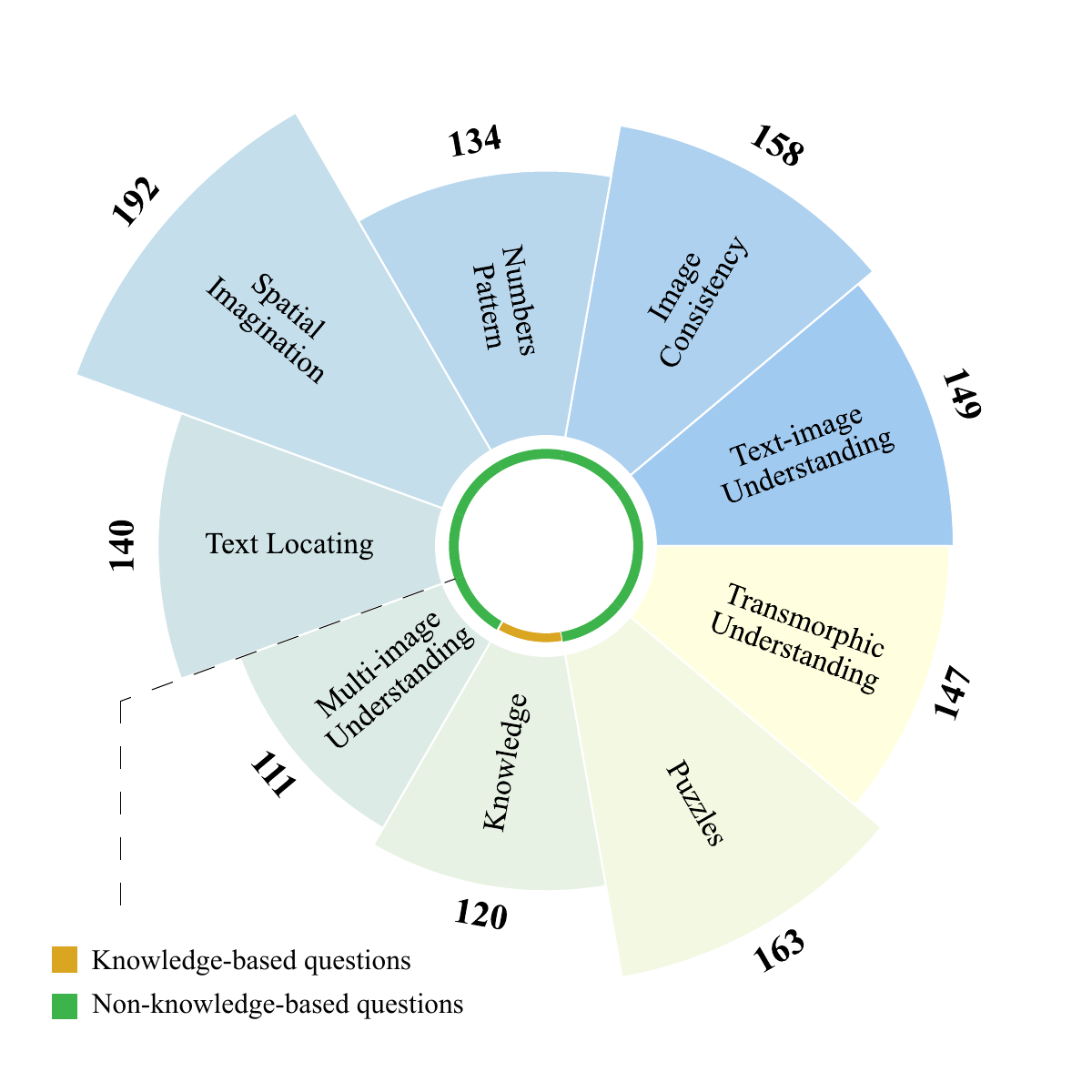}
    \caption{Statistics of MANBench. The benchmark comprises 9 tasks, categorized into knowledge-based and non-knowledge-based questions.}
    \label{fig:manbench}
  \end{figure}
  \begin{figure}[h]
    \centering
    \includegraphics[width=0.8\columnwidth]{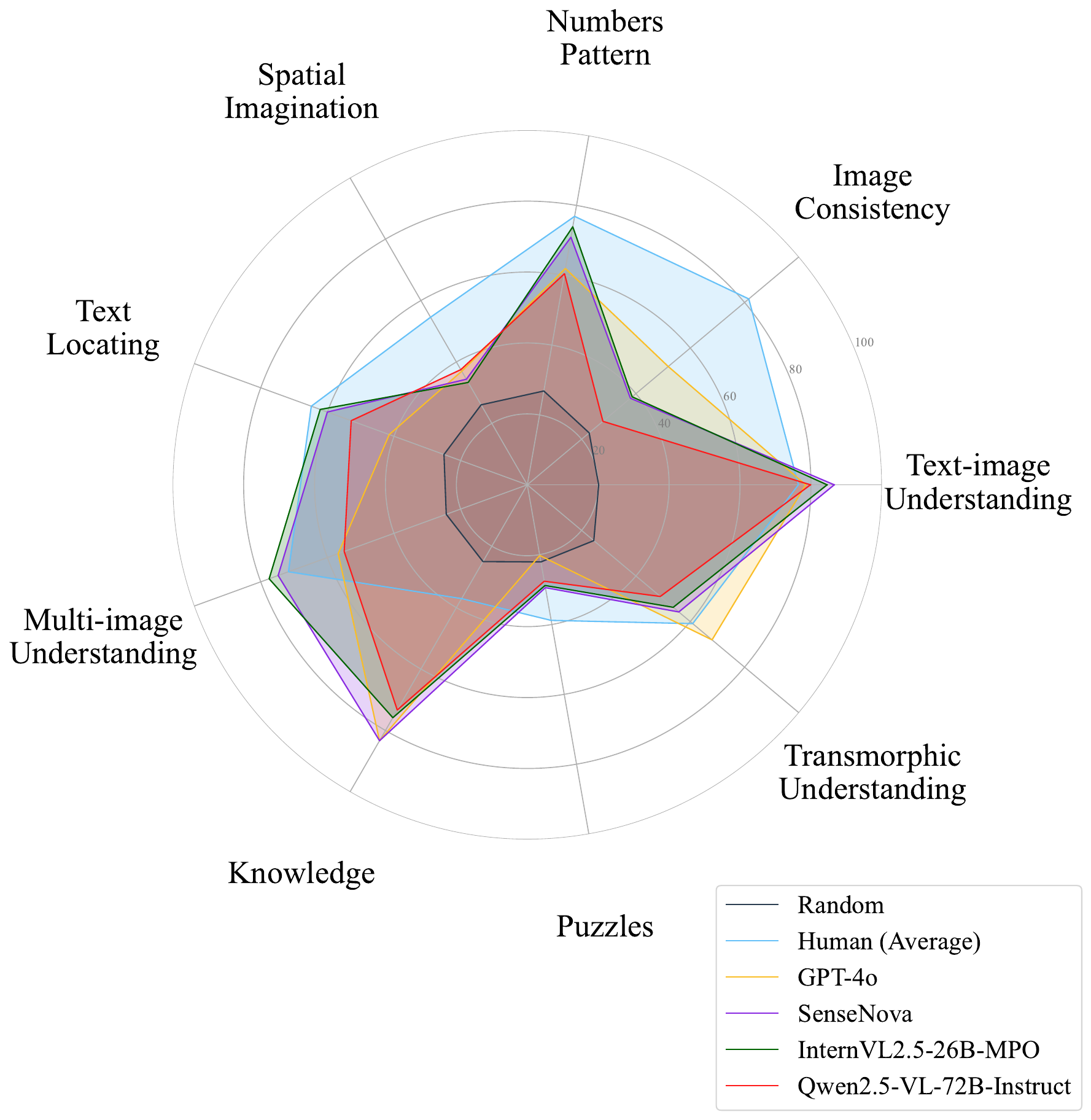}
    \caption{Performance comparison on the MANBench English subset among human average accuracy and some selected MLLMs. Please refer to Table~\ref{tab:performance} for more results.}
    \label{fig:radar} 
  \end{figure}

As illustrated in Figure~\ref{fig:manbench}, MANBench comprises nine tasks, each containing more than 110 questions, resulting in a total of 1,314 questions and 2,231 images.
All questions in the dataset are bilingual, with text provided in both English and Chinese. The translations have been meticulously reviewed by the coauthors to ensure semantic equivalence across languages.

The questions are categorized into two types: knowledge-based and non-knowledge-based. Knowledge-based questions require participants to possess prior knowledge in subjects such as Biology, Engineering, Geography, Physics, and Chemistry. In contrast, non-knowledge-based questions do not demand advanced domain-specific knowledge.

\subsection{Key Features of MANBench}
\textbf{Intuitive Reasoning without Domain Bias:} The benchmark is designed to eliminate reliance on prior knowledge by emphasizing universal perception and logical inference, accessible through immediate visual-textual analysis. The tasks are carefully constructed to avoid domain-specific knowledge, ensuring ease of use and fairness in benchmarking.

\noindent\textbf{Seamless Cross-Modal Integration:} Questions necessitate joint interpretation of images and text, demanding causal or functional relationships between modalities rather than superficial keyword-to-image alignment. This design compels subjects to synthesize contextual meaning from both inputs, such as inferring intent from visual details paired with textual hints.

\noindent\textbf{Clarity-Focused Task Design:} The tasks prioritize clear and unambiguous question stems with minimal distractors, ensuring that the challenges arise from reasoning difficulty rather than linguistic ambiguity or misleading cues.

\noindent\textbf{Robustness to Real-World Complexity:} By simulating high-density, noisy environments and requiring multi-step reasoning, the benchmark discourages shortcut solutions. This ensures that models demonstrate practical applicability in solving complex, real-world scenarios.

\subsection{Dataset Collection Process}
MANBench evaluates multimodal capabilities across nine tasks, all of which utilize a multiple-choice question format. The dataset is constructed from diverse sources, including generated images, real-world images, social media images, and images from existing benchmark datasets.

\noindent\textbf{Spatial Imagination:} This task systematically evaluates the spatial cognition capabilities of MLLMs and humans through a multidimensional assessment framework. By utilizing generated controllable scene images, real-world images, and 3D IQ test images, we design a variety of question types, including 2D-to-3D spatial transformations, the disassembly and reassembly of three-dimensional structures, spatial distance measurements, and object quantity estimations. Through the simulation of real-world scenarios and abstract geometric problems, the task comprehensively examines subjects’ mastery of core competencies, such as spatial topological relationships and geometric transformation principles.

\noindent\textbf{Numbers Pattern:} This task is designed to comprehensively evaluate the numerical reasoning capabilities of MLLMs and humans. By integrating generated original images with real-world photographic materials, we design question types that span precise object counting, dynamic quantity estimation, and complex numerical pattern deduction. These multidimensional questions systematically assess the ability of models to process numerical information, ranging from foundational concepts to advanced reasoning.

\noindent\textbf{Puzzles:} This task aims to evaluate the inductive and reasoning abilities of MLLMs and humans in abstract graphic patterns. The data sources for this task are diverse and unique: On the one hand, it utilizes visual classic IQ test~\citep{fu2024blink}, presenting problems through basic shape examples and four candidate images to assess subjects’ logical reasoning capabilities based on shape attributes, positional relationships, and sequential patterns. On the other hand, we develop maze schematics to evaluate subjects’ ability to estimate path distances in planar mazes. As a type of problem rarely addressed in existing datasets, this task provides a novel means of assessing the ability to extract and reason about information from complex graphics.

\noindent\textbf{Multi-image Understanding:} This task aims to assess the ability of MLLMs and humans to synthesize and interpret information from multiple images. We collect a diverse set of images from the Internet and MUIRBench~\citep{wang2024muirbenchcomprehensivebenchmarkrobust}, as well as generate additional images, carefully constructing a series of questions that require models to accurately identify commonalities among multiple images. These questions require the model not only to comprehend the content of each image individually but also to compare and reason across multiple images to discern shared features or patterns. 

\noindent\textbf{Text-image Understanding:} This task evaluates the ability of MLLMs to comprehensively integrate and understand both image and text information. Unlike traditional visual quizzes that rely on superficial descriptions, this task emphasizes three specific types of questions: phenomenon attribution, functional reasoning about objects, and topic generalization. The dataset for this task is constructed by integrating resources from MMBench~\citep{MMbench}, MMStar~\citep{chen2024we}, and social media.

\noindent\textbf{Image Consistency:} The task is designed to assess the ability of MLLMs and humans to understand the consistency of an image locally and holistically. By integrating existing benchmarks, social media, and generated visuals, we implement a non-uniform grid partitioning strategy. Specifically, fixed-size cropping is applied to the bottom-right corner of source images, while the remaining area is retained as a reference. Subjects are required to identify the visually consistent completion unit from a set of four candidate images. Compared to the jigsaw task of BLINK~\citep{fu2024blink}, our design improves evaluation precision for cross-scale visual-semantic alignment and detail-oriented reasoning by expanding candidate options and incorporating more complex visual patterns.

\noindent\textbf{Transmorphic Understanding:} The aim of this task is to evaluate the generalization ability of MLLMs and humans to trigger cross-modal abstract associations from visual inputs and overcome the limitations of figurative semantic matching. For this purpose, we collect images from social media and require subjects to infer underlying human emotions based on individual images or extrapolate from the visual characteristics of objects to implicitly associated scenes.

\noindent\textbf{Text Locating:} This task aims to evaluate the ability of subjects to accurately locate targets in images based on textual descriptions, which is crucial for multimodal interaction and image-text fusion applications. To construct the task, we collect images from social media and annotate them with multiple markers, accompanied by a textual description. These markers follow the design principles of BLINK~\citep{fu2024blink}. Then, subjects are required to select the target point based on the description. Unlike traditional tests that focus primarily on basic image-text matching, this task emphasizes a deeper comprehension of textual information and efficient retrieval capabilities in complex scenarios, thereby offering more comprehensive and precise evaluation results.

\noindent\textbf{Knowledge:} This task is designed to assess the knowledge base, conceptual associations, and cross-modal knowledge transfer in humans and MLLMs. With data from Wikipedia, OxfordPets~\citep{oxfordpets}, ScienceQA~\citep{lu2022learn} and MMStar~\citep{chen2024we}, we generate a series of questions that require subjects to recognise images, reason about them according to their knowledge system, and select the matching textual options.

\section{Experiments}
We conducted extensive experiments involving 575 human participants and 12 MLLMs to evaluate their multimodal abilities. 
\subsection{Experimental Setup}
\subsubsection{Human Experiments}
The human experiments were conducted in two stages. 

In the first stage, a preliminary pilot study was conducted to evaluate the quality of the questions. Five native Chinese-speaking university students were recruited to assess the questions. They completed all tasks in the dataset and provided feedback on clarity, ambiguity, and difficulty. Their response times and accuracy were analyzed to determine the difficulty level of the questions.

In the second stage of our study, we expanded our participant pool to include a larger and more diverse group. Given the impracticality of requiring participants to complete the entire dataset of 1,314 questions in a large-scale study—which would take approximately 10 hours—we divided the dataset into ten subsets, each containing around 131 questions. This division was guided by a difficulty gradient established during the pilot study. However, the limited number of participants in the pilot study may have introduced some imperfections in the gradient. Despite this limitation, the division facilitated a more manageable and scalable evaluation process for our broader study.

\begin{figure}[h]
    \centering
    \includegraphics[width=0.9\columnwidth]{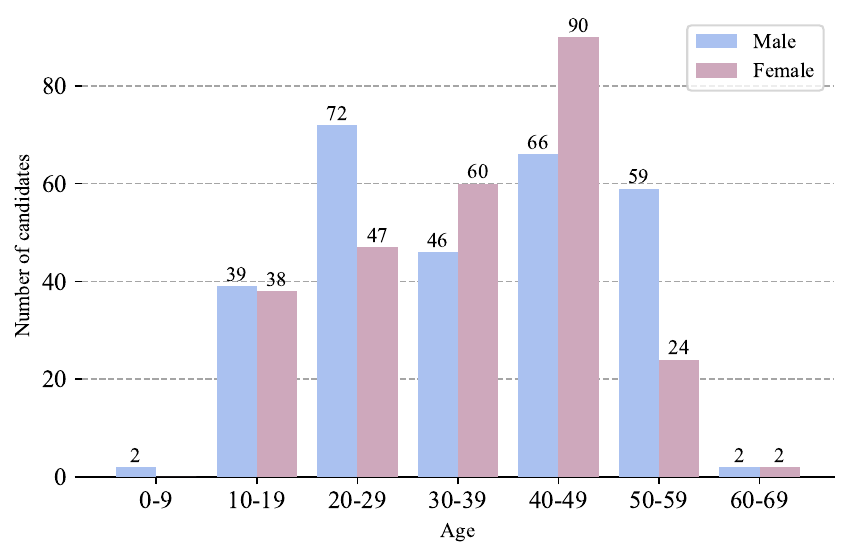}
    \caption{Age and gender distribution of participants.}
    \label{fig:age_gender_distribution}
\end{figure}

We then recruit 575 Chinese participants from diverse backgrounds to engage in the experiments. Additionally, we gather demographic data from the participants, including age, gender, and education level, to enhance the analysis. As shown in Figure~\ref{fig:age_gender_distribution}, the participants are distributed across a wide range of ages. For more participants statistics, please refer to Appendix~\ref{sec:human_experiments_statistics}.

We meticulously designed the user interface for the human experiments to ensure clarity and consistency, allowing participants to clearly comprehend the questions, provide responses, and receive the same information as the MLLMs. Participants were permitted to omit perceived as overly challenging or ambiguous. For further details regarding the experimental setup, methodology, and additional resources, please refer to Appendix~\ref{sec:human_experiments}.

\subsection{MLLM Experiments}
We adopt standard experimental setups following the VLMEvalKit~\citep{duan2024vlmevalkit}, where the temperature is fixed at 0 and the retry count is set to 10. Please refer to Appendix~\ref{sec:mlm_experiments} for more details on our computing infrastructure.

The MANBench benchmark is evaluated on 14 recent MLLMs. Sepecifically, we select 6 closed-source models: GPT-4o~\citep{gpt4o}, GPT-o1~\citep{gpto1}, SenseNova~\citep{SenseNova}, Step-1o~\citep{stepfun}, Gemini-1.5-Pro~\citep{team2023gemini}, Claude-3.5-Sonnet~\citep{claude}. Additionally, we evaluate 8 open-source models: Deepseek-VL2~\citep{wu2024deepseekvl2mixtureofexpertsvisionlanguagemodels}, Qwen2-VL-72B-Instruct~\citep{Qwen2VL}, Qwen2.5-VL-72B~\citep{qwen2.5-VL}, QVQ-72B-Preview~\citep{qvq-72b-preview}, InternVL2-26B, InternVL2-8B~\citep{chen2024internvl}, InternVL2.5-26B-MPO, InternVL2.5-78B-MPO~\citep{wang2024mpo}.

\subsection{Main Results}
\begin{table*}
  \centering
\resizebox{\textwidth}{!}{
  \begin{tabular}{lcccccccccc}
    \toprule[1.2pt]
   & \textbf{\scriptsize{\makecell{Overall}}} & \textbf{\scriptsize{\makecell{Text-image \\ Understanding \\ (149)}}} & \textbf{\scriptsize{\makecell{Numbers \\ Pattern \\ (134)}}} & \textbf{\scriptsize{\makecell{Image \\ Consistency \\ (158)}}} & \textbf{\scriptsize{\makecell{Spatial \\ Imagination \\ (192)}}} & \textbf{\scriptsize{\makecell{Text \\ Locating \\ (140)}}} & \textbf{\scriptsize{\makecell{Multi-image \\ Understanding \\ (111)}}} & \textbf{\scriptsize{\makecell{Knowledge \\ (120)}}} & \textbf{\scriptsize{\makecell{Puzzles \\ (163)}}} & \textbf{\scriptsize{\makecell{Transmorphic \\ Understanding \\ (147)}}} \\
   \midrule[1.2pt]
    Human (Average) & 62.26 & 76.46 & 76.88 & 81.55 & 54.56 & 64.82 & 71.69 & 37.05 & 38.83 & 60.90 \\
    Human (Best) & 90.87 & 89.00 & 91.79 & 100.00 & 94.57 & 94.39 & 92.86 & 74.75 & 86.42 & 90.54 \\
    Random & 24.05 & 20.13 & 26.87 & 22.78 & 26.04 & 25.00 & 24.32 & 25.00 & 22.09 & 24.49 \\
\hline \multicolumn{11}{c}{ \textbf{Open-source  MLLMs}} \\ \hline 
    Deepseek-VL2 & 45.43 & \underline{76.51} & 57.46 & 31.01 & 31.77 & 44.29 & 36.04 & \underline{70.83} & 26.38 & 44.90 \\
    InternVL2-26B & 41.86 & 61.74 & 54.48 & 24.68 & 29.69 & 33.57 & 43.24 & \underline{67.50} & 27.61 & 46.26 \\
    InternVL2-8B & 36.15 & 59.06 & 36.57 & 27.22 & 27.08 & 31.43 & 31.53 & \underline{50.83} & 28.22 & 38.78 \\
    InternVL2.5-26B-MPO & 56.32 & \textbf{\underline{84.56}} & \textbf{73.88} & 38.61 & 33.33 & 62.14 & \textbf{\underline{77.48}} & \underline{75.83} & 28.83 & 53.74 \\
    InternVL2.5-78B-MPO & \textbf{59.82} & \textbf{\underline{84.56}} & 70.90 & \textbf{45.57} & \textbf{38.02} & \textbf{62.86} & \underline{75.68} & \underline{80.00} & \textbf{38.65} & \textbf{60.54} \\
    QVQ-72B-Preview & 50.00 & \underline{76.51} & 58.21 & 31.01 & 31.77 & 42.86 & 63.96 & \textbf{\underline{82.50}} & 34.36 & 46.94 \\
    Qwen2-VL-72B-Instruct & 46.19 & \underline{77.85} & 58.96 & 28.48 & 32.29 & 38.57 & 48.65 & \underline{79.17} & 23.31 & 43.54 \\
    Qwen2.5-VL-72B-Instruct & 49.92 & \underline{79.87} & 60.45 & 27.85 & 37.50 & 52.86 & 54.95 & \underline{73.33} & 27.61 & 48.98 \\
\hline \multicolumn{11}{c}{ \textbf{Closed-source MLLMs}} \\ \hline 
    Claude-3.5-Sonnet & 54.87 & \underline{81.88} & 57.46 & 46.20 & 36.98 & 50.00 & 69.37 & \underline{78.33} & 27.61 & \underline{62.59} \\
    GPT-4o & 53.81 & \underline{78.52} & 61.94 & 51.90 & 36.98 & 41.43 & 56.76 & \underline{83.33} & 20.25 & \underline{68.03} \\
    GPT-o1 & \textbf{59.97} & \textbf{\underline{88.59}} & \textbf{73.88} & \textbf{53.16} & 34.90 & 54.29 & 64.86 & \textbf{\underline{85.83}} & 33.74 & \underline{68.03} \\
    Gemini-1.5-Pro & 55.10 & \underline{86.58} & 66.42 & 33.54 & \textbf{41.67} & 39.29 & 69.37 & \underline{83.33} & \textbf{34.36} & 57.82 \\
    SenseNova & 56.85 & \underline{86.58} & 70.90 & 37.97 & 34.38 & \textbf{60.00} & \textbf{\underline{74.77}} & \underline{83.33} & 29.45 & 55.78 \\
    Step-1o & 54.79 & \underline{83.89} & 68.66 & 39.87 & 30.73 & 44.29 & 60.36 & \underline{78.33} & 33.74 & \textbf{\underline{70.07}} \\
     \bottomrule
  \end{tabular}
}
\caption{Results of different models on the MANBench English subset. The first row shows task names and number of test data. The best performance in each task is in bold. Values exceeding human average performance are underlined. For detailed results of Chinese subset, please refer to Table~\ref{tab:performance_chinese}. For the information of the best human performance, please refer to Appendix~\ref{sec:best_human_performance}. The overall score is calculated by weighting the task scores according to the number of questions in each task.}
\label{tab:performance}
\end{table*}

\subsubsection{MLLMs Performance}
The overall performance of MLLMs on MANBench is shown in Table~\ref{tab:performance}. The best MLLMs performance is achieved by the closed-source model GPT-o1 with the average score of 59.97, followed by the open-source model InternVL2.5-78B-MPO with a score of 59.82. 

Especially on the Puzzles task,  the accuracy of MLLMs is approximately 30\%, which is comparable to random guessing. This result highlights the significant difficulty of the Puzzles task for MLLMs and suggests that there remains substantial room for improvement in this type of image reasoning task.

\subsubsection{Human Performance}
We evaluate the performance of MLLMs by comparing their results to human performance benchmarks. As detailed in Table~\ref{tab:performance}, two key metrics are employed: average human performance and best human performance. The average human performance serves as the baseline for assessing MLLM capabilities. To compute this baseline, the accuracy for each task \( C_i \) is determined as follows: For every question \( j \) within a task, the accuracy \( p_{ij} \) is calculated. The task accuracy \( P_i \) is then derived by averaging the accuracies of all questions within the task:
\[
P_i = \frac{1}{n_i} \sum_{j=1}^{n_i} p_{ij}
\]
where \( n_i \) represents the total number of questions in task \( C_i \).

\begin{figure}[h]
    \centering
    \includegraphics[width=0.8\columnwidth]{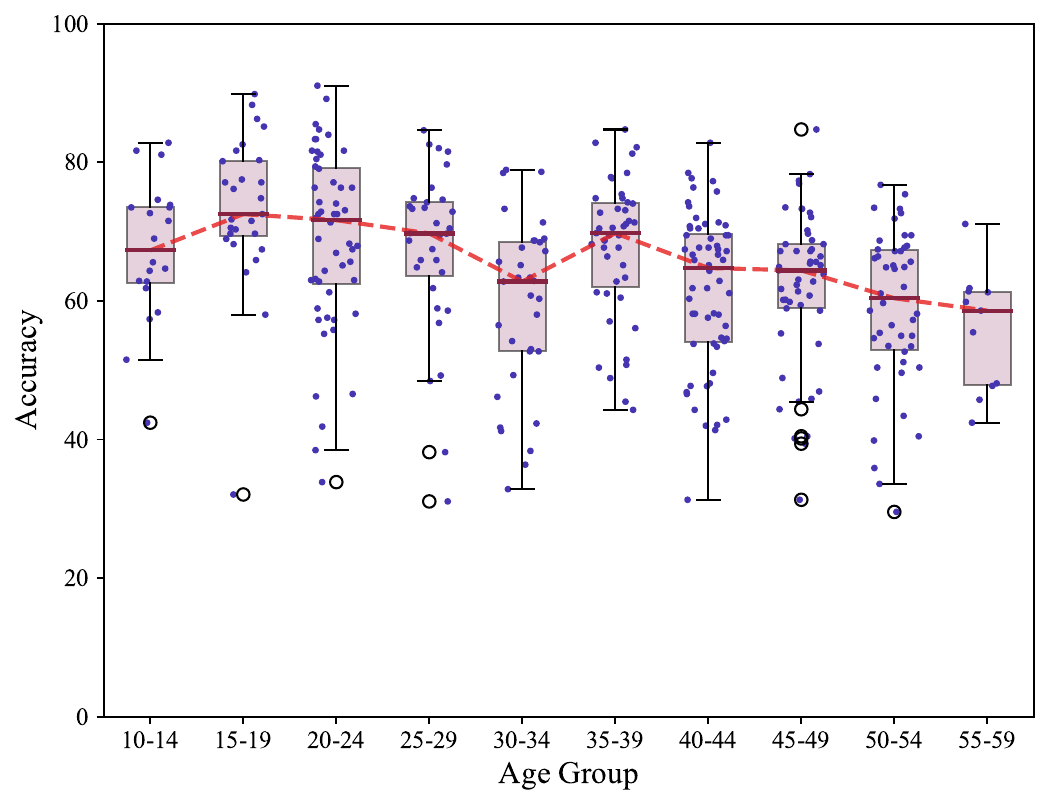}
    \caption{Performance comparison across different age groups.}
    \label{fig:age_performance}
\end{figure}

In Figure~\ref{fig:age_performance}, we observe that children under the age of fourteen perform comparably to adults on MANBench, indicating that our dataset is not highly sensitive to age differences. Nevertheless, our findings also reveal a decline in human performance as age increases.

\subsection{Analysis}
Analysis of Table~\ref{tab:performance} shows humans generally outperform MLLMs across most tasks, except in Knowledge and Text-Image Understanding. This suggests that MLLMs excel in tasks requiring prior knowledge and basic text-image alignment but struggle in tasks demanding deeper cross-modal reasoning. To further investigate the factors influencing task performance, we conducted additional analyses to address three key questions: the role of prior knowledge, the necessity of visual content, and the sufficiency of reasoning difficulty.

\noindent\textbf{Is prior knowledge crucial?}
As shown in Figure~\ref{fig:age_performance}, children under the age of fourteen perform comparably to adults on MANBench, despite likely possessing less prior knowledge. This indicates that prior knowledge is not crucial for task performance in MANBench, emphasizing the dataset’s focus on reasoning rather than domain-specific expertise.

\noindent\textbf{Is the visual contents necessary?} 
\begin{figure}
    \centering
    \includegraphics[width=0.95\columnwidth]{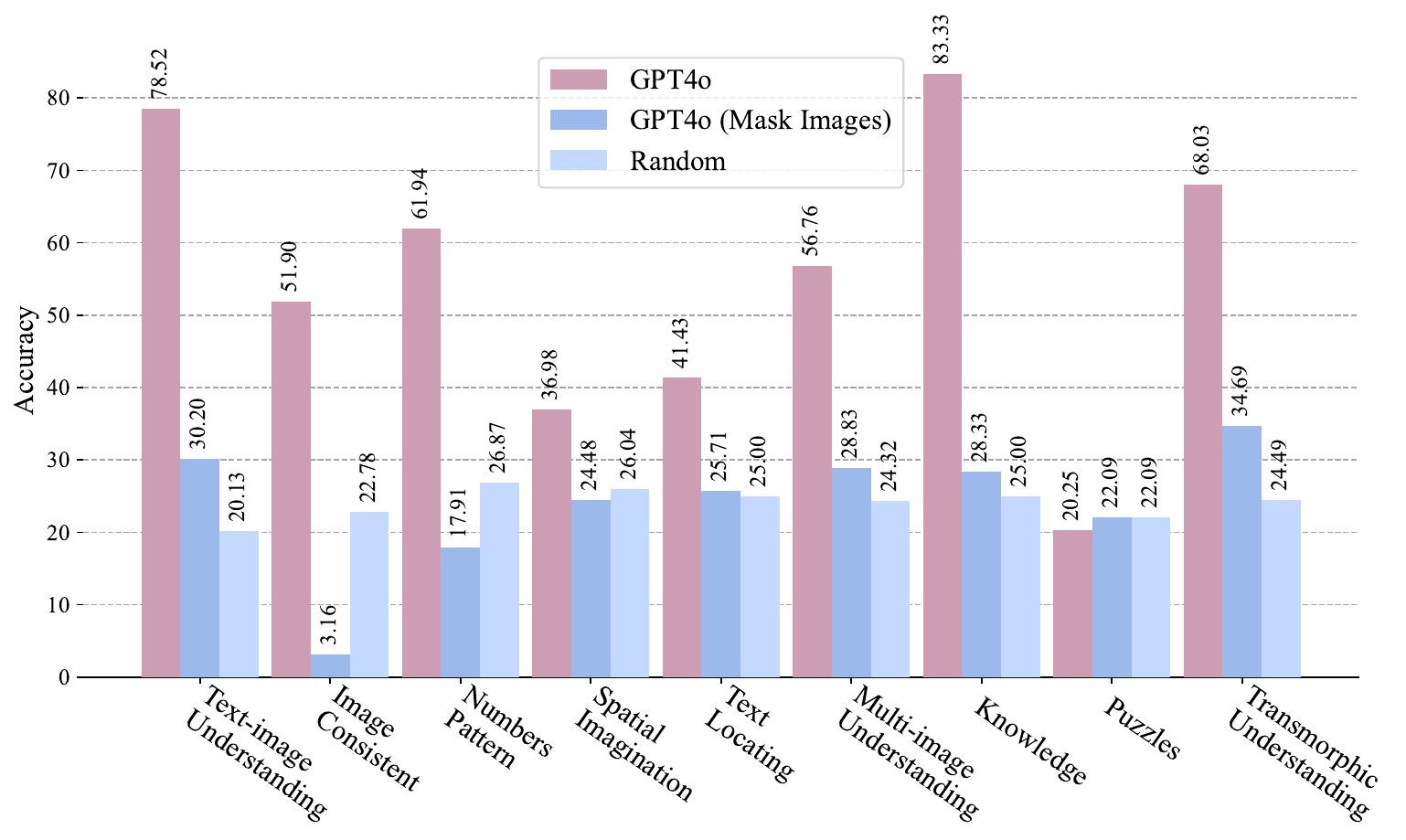}
    \caption{Performance of GPT-4o with and without images on MANBench English subset.}
    \label{fig:ablation}
\end{figure}
To investigate the necessity of visual content in the tasks, we conducted an ablation study on GPT-4o by removing the images from the questions. For more experiment details, please refer to Appendix~\ref{sec:ablation_study}.  As shown in Figure~\ref{fig:ablation}, GPT-4o demonstrated performance divergence across 9 task categories. In the Image Consistency task, due to the lack of images, GPT-4o refused to answer most questions, which confirmed strong visual dependency. And in the Transmorphic Understanding task, the model’s accuracy slightly exceeded random baseline. A further manual review revealed that this higher accuracy resulted from the model’s preference for positive emotions, such as “joy”, which appear more frequently as correct answers in the task. For other task types, the model’s performance showed no statistically significant deviation from random chance.

\begin{figure}
    \centering
    \includegraphics[width=\columnwidth]{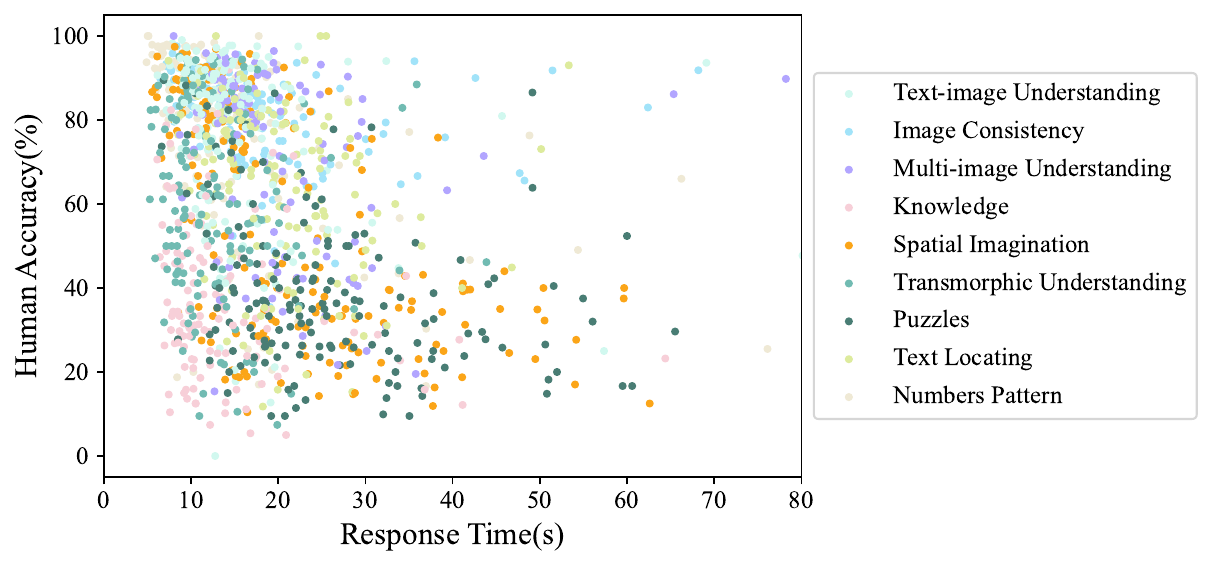}
    \caption{Average human response time for all questions in MANBench. Different colors indicate distinct categories within MANBench. Each point represents the average response time and accuracy for a specific question.}
    \label{fig:response_time}
\end{figure}

\noindent\textbf{Is the reasoning difficulty sufficient?}
To assess the reasoning difficulty of the tasks, we measured the response time of participants for each question. The results, presented in Figure~\ref{fig:response_time}, reveal a wide range of reasoning response times among humans on MANBench, which correlates with varying levels of accuracy. This indicates that MANBench encompasses a diverse spectrum of question difficulties, effectively evaluating the reasoning capabilities of both humans and MLLMs.
\section{Conclusion}
The advancement of artificial intelligence reflects humanity’s enduring pursuit to replicate and understand the complexities of human cognition. MANBench, as a meticulously designed benchmark, contributes to this endeavor by evaluating the multimodal capabilities of both humans and MLLMs. Our experiments reveal that while MLLMs excel in tasks such as text generation and knowledge retrieval, they encounter significant challenges in tasks requiring nuanced cross-modal reasoning. Even state-of-the-art models, including GPT-4o, Qwen2.5-VL-72B-Instruct, and Gemini-1.5 Pro, achieve less than 60\% accuracy on MANBench, falling short of average human performance and remaining far from the upper bounds of human capability.

MANBench provides a structured framework for assessing the strengths and limitations of MLLMs, offering valuable insights into areas requiring further development. It highlights the importance of moving beyond superficial alignment toward deeper, integrative multimodal reasoning. While MANBench is a step forward, it is not without limitations, and its findings should be interpreted as part of a broader effort to advance AI. As the field progresses, benchmarks like MANBench can serve as tools to guide research and inspire the development of systems that better approximate human understanding. Ultimately, the goal is not merely to surpass benchmarks but to foster meaningful progress in bridging the gap between artificial and human intelligence.

\section{Limitations}
\noindent\textbf{Low population coverage:} The experimental subjects in our study were predominantly drawn from a specific group within the Chinese linguistic and cultural context, which may restrict the cross-linguistic and cross-cultural adaptability of our findings. Cultural differences, variations in knowledge bases, and distinct cognitive patterns could significantly impact the generalizability of the results. Additionally, the educational backgrounds and age distributions of our participants do not fully represent the broader human population, potentially limiting the generalizability of the human benchmarks established in this study. Please refer to Appendix~\ref{sec:human_experiments_statistics} for more details on participant demographics.

\noindent\textbf{Image Attribution and Copyright Compliance:} The images utilized in our dataset were obtained from publicly available online sources. We have diligently ensured that all images included in this paper comply with applicable copyright laws and are accompanied by proper attribution. However, if you are the copyright holder of any image featured in our work and believe its usage violates your licensing terms, please contact us immediately. We are committed to resolving any legitimate copyright concerns in a timely and respectful manner. Please refer to Appendix~\ref{sec:artifact_licensing} for more details on data licensing.
\section{Ethical Considerations}
\noindent\textbf{Informed Consent:} Prior to submitting personal information through the web interface, all participants were fully informed about the purpose of data collection and its exclusive use for scientific research.

\noindent\textbf{Protection of Minors:}
We ensured that all minor participants completed the survey under the supervision of a guardian.

\noindent\textbf{Privacy Protection:}
Participant privacy was strictly protected, with no personally identifiable information being analyzed or disclosed during the data analysis process. Appendix~\ref{sec:best_human_performance} was written only after obtaining the explicit consent of the participant involved.

\noindent\textbf{Risk of Amplifying Educational Bias:} The overrepresentation of highly educated participants (e.g., 40.5\% with bachelor’s degrees) in the dataset may inadvertently reinforce existing societal biases. If misused, such data could deepen disparities in education and employment by framing high academic achievement as a universal competency standard.

\bibliography{custom}

\begin{thebibliography}{38}
\providecommand{\natexlab}[1]{#1}

\bibitem[{Amini et~al.(2019)Amini, Gabriel, Lin, Koncel-Kedziorski, Choi, and Hajishirzi}]{amini2019mathqa}
Aida Amini, Saadia Gabriel, Peter Lin, Rik Koncel-Kedziorski, Yejin Choi, and Hannaneh Hajishirzi. 2019.
\newblock Mathqa: Towards interpretable math word problem solving with operation-based formalisms.
\newblock \emph{arXiv preprint arXiv:1905.13319}.

\bibitem[{Anthropic(2024)}]{claude}
Anthropic. 2024.
\newblock Claude 3.5 sonnet.
\newblock \url{https://www.anthropic.com/news/claude-3-5-sonnet}.

\bibitem[{Chen et~al.(2024{\natexlab{a}})Chen, Li, Dong, Zhang, Zang, Chen, Duan, Wang, Qiao, Lin et~al.}]{chen2024we}
Lin Chen, Jinsong Li, Xiaoyi Dong, Pan Zhang, Yuhang Zang, Zehui Chen, Haodong Duan, Jiaqi Wang, Yu~Qiao, Dahua Lin, et~al. 2024{\natexlab{a}}.
\newblock Are we on the right way for evaluating large vision-language models?
\newblock \emph{arXiv preprint arXiv:2403.20330}.

\bibitem[{Chen et~al.(2024{\natexlab{b}})Chen, Ye, Wang, Li, Deng, Li, Li, Duan, Huang, Su et~al.}]{chen2024gmai}
Pengcheng Chen, Jin Ye, Guoan Wang, Yanjun Li, Zhongying Deng, Wei Li, Tianbin Li, Haodong Duan, Ziyan Huang, Yanzhou Su, et~al. 2024{\natexlab{b}}.
\newblock Gmai-mmbench: A comprehensive multimodal evaluation benchmark towards general medical ai.
\newblock \emph{arXiv preprint arXiv:2408.03361}.

\bibitem[{Chen et~al.(2024{\natexlab{c}})Chen, Wu, Wang, Su, Chen, Xing, Zhong, Zhang, Zhu, Lu et~al.}]{chen2024internvl}
Zhe Chen, Jiannan Wu, Wenhai Wang, Weijie Su, Guo Chen, Sen Xing, Muyan Zhong, Qinglong Zhang, Xizhou Zhu, Lewei Lu, et~al. 2024{\natexlab{c}}.
\newblock Internvl: Scaling up vision foundation models and aligning for generic visual-linguistic tasks.
\newblock In \emph{Proceedings of the IEEE/CVF Conference on Computer Vision and Pattern Recognition}, pages 24185--24198.

\bibitem[{Duan et~al.(2024)Duan, Yang, Qiao, Fang, Chen, Liu, Dong, Zang, Zhang, Wang et~al.}]{duan2024vlmevalkit}
Haodong Duan, Junming Yang, Yuxuan Qiao, Xinyu Fang, Lin Chen, Yuan Liu, Xiaoyi Dong, Yuhang Zang, Pan Zhang, Jiaqi Wang, et~al. 2024.
\newblock Vlmevalkit: An open-source toolkit for evaluating large multi-modality models.
\newblock In \emph{Proceedings of the 32nd ACM International Conference on Multimedia}, pages 11198--11201.

\bibitem[{Dubey et~al.(2024)Dubey, Jauhri, Pandey, Kadian, Al-Dahle, Letman, Mathur, Schelten, Yang, Fan et~al.}]{dubey2024llama}
Abhimanyu Dubey, Abhinav Jauhri, Abhinav Pandey, Abhishek Kadian, Ahmad Al-Dahle, Aiesha Letman, Akhil Mathur, Alan Schelten, Amy Yang, Angela Fan, et~al. 2024.
\newblock The llama 3 herd of models.
\newblock \emph{arXiv preprint arXiv:2407.21783}.

\bibitem[{Fu et~al.(2024)Fu, Hu, Li, Feng, Wang, Lin, Roth, Smith, Ma, and Krishna}]{fu2024blink}
Xingyu Fu, Yushi Hu, Bangzheng Li, Yu~Feng, Haoyu Wang, Xudong Lin, Dan Roth, Noah~A Smith, Wei-Chiu Ma, and Ranjay Krishna. 2024.
\newblock Blink: Multimodal large language models can see but not perceive.
\newblock \emph{arXiv preprint arXiv:2404.12390}.

\bibitem[{Jaech et~al.(2024)Jaech, Kalai, Lerer, Richardson, El-Kishky, Low, Helyar, Madry, Beutel, Carney et~al.}]{jaech2024openai}
Aaron Jaech, Adam Kalai, Adam Lerer, Adam Richardson, Ahmed El-Kishky, Aiden Low, Alec Helyar, Aleksander Madry, Alex Beutel, Alex Carney, et~al. 2024.
\newblock Openai o1 system card.
\newblock \emph{arXiv preprint arXiv:2412.16720}.

\bibitem[{Liu et~al.(2024{\natexlab{a}})Liu, Feng, Xue, Wang, Wu, Lu, Zhao, Deng, Zhang, Ruan et~al.}]{liu2024deepseek}
Aixin Liu, Bei Feng, Bing Xue, Bingxuan Wang, Bochao Wu, Chengda Lu, Chenggang Zhao, Chengqi Deng, Chenyu Zhang, Chong Ruan, et~al. 2024{\natexlab{a}}.
\newblock Deepseek-v3 technical report.
\newblock \emph{arXiv preprint arXiv:2412.19437}.

\bibitem[{Liu et~al.(2024{\natexlab{b}})Liu, Li, Li, and Lee}]{liu2024improved}
Haotian Liu, Chunyuan Li, Yuheng Li, and Yong~Jae Lee. 2024{\natexlab{b}}.
\newblock Improved baselines with visual instruction tuning.
\newblock In \emph{Proceedings of the IEEE/CVF Conference on Computer Vision and Pattern Recognition}, pages 26296--26306.

\bibitem[{Liu et~al.(2024{\natexlab{c}})Liu, Li, Li, Li, Zhang, Shen, and Lee}]{liu2024llava}
Haotian Liu, Chunyuan Li, Yuheng Li, Bo~Li, Yuanhan Zhang, Sheng Shen, and Yong~Jae Lee. 2024{\natexlab{c}}.
\newblock Llava-next: Improved reasoning, ocr, and world knowledge.

\bibitem[{Liu et~al.(2024{\natexlab{d}})Liu, Duan, Zhang, Li, Zhang, Zhao, Yuan, Wang, He, Liu, Chen, and Lin}]{MMbench}
Yuan Liu, Haodong Duan, Yuanhan Zhang, Bo~Li, Songyang Zhang, Wangbo Zhao, Yike Yuan, Jiaqi Wang, Conghui He, Ziwei Liu, Kai Chen, and Dahua Lin. 2024{\natexlab{d}}.
\newblock \href {https://arxiv.org/abs/2307.06281} {Mmbench: Is your multi-modal model an all-around player?}
\newblock \emph{Preprint}, arXiv:2307.06281.

\bibitem[{Lu et~al.(2024)Lu, Bansal, Xia, Liu, Li, Hajishirzi, Cheng, Chang, Galley, and Gao}]{lu2024mathvista}
Pan Lu, Hritik Bansal, Tony Xia, Jiacheng Liu, Chunyuan Li, Hannaneh Hajishirzi, Hao Cheng, Kai-Wei Chang, Michel Galley, and Jianfeng Gao. 2024.
\newblock Mathvista: Evaluating mathematical reasoning of foundation models in visual contexts.
\newblock In \emph{International Conference on Learning Representations (ICLR)}.

\bibitem[{Lu et~al.(2022)Lu, Mishra, Xia, Qiu, Chang, Zhu, Tafjord, Clark, and Kalyan}]{lu2022learn}
Pan Lu, Swaroop Mishra, Tony Xia, Liang Qiu, Kai-Wei Chang, Song-Chun Zhu, Oyvind Tafjord, Peter Clark, and Ashwin Kalyan. 2022.
\newblock Learn to explain: Multimodal reasoning via thought chains for science question answering.
\newblock In \emph{The 36th Conference on Neural Information Processing Systems (NeurIPS)}.

\bibitem[{Meng et~al.(2024)Meng, Wang, Li, Lu, Tian, Liao, Zhu, Dai, Qiao, Luo et~al.}]{meng2024mmiu}
Fanqing Meng, Jin Wang, Chuanhao Li, Quanfeng Lu, Hao Tian, Jiaqi Liao, Xizhou Zhu, Jifeng Dai, Yu~Qiao, Ping Luo, et~al. 2024.
\newblock Mmiu: Multimodal multi-image understanding for evaluating large vision-language models.
\newblock \emph{arXiv preprint arXiv:2408.02718}.

\bibitem[{OpenAI(2024{\natexlab{a}})}]{gpto1}
OpenAI. 2024{\natexlab{a}}.
\newblock Gpt-4o.
\newblock \url{https://openai.com/index/hello-gpt-4o/}.

\bibitem[{OpenAI(2024{\natexlab{b}})}]{gpt4o}
OpenAI. 2024{\natexlab{b}}.
\newblock Gpt-o1.
\newblock \url{https://openai.com/index/openai-o1-system-card/}.

\bibitem[{Parkhi et~al.(2012)Parkhi, Vedaldi, Zisserman, and Jawahar}]{oxfordpets}
Omkar~M Parkhi, Andrea Vedaldi, Andrew Zisserman, and C.~V. Jawahar. 2012.
\newblock \href {https://doi.org/10.1109/CVPR.2012.6248092} {Cats and dogs}.
\newblock In \emph{2012 IEEE Conference on Computer Vision and Pattern Recognition}, pages 3498--3505.

\bibitem[{Radford et~al.(2021)Radford, Kim, Hallacy, Ramesh, Goh, Agarwal, Sastry, Askell, Mishkin, Clark et~al.}]{radford2021learning}
Alec Radford, Jong~Wook Kim, Chris Hallacy, Aditya Ramesh, Gabriel Goh, Sandhini Agarwal, Girish Sastry, Amanda Askell, Pamela Mishkin, Jack Clark, et~al. 2021.
\newblock Learning transferable visual models from natural language supervision.
\newblock In \emph{International conference on machine learning}, pages 8748--8763. PMLR.

\bibitem[{Russakovsky et~al.(2014)Russakovsky, Deng, Su, Krause, Satheesh, Ma, Huang, Karpathy, Khosla, Bernstein, Berg, and Fei-Fei}]{ImageNet}
Olga Russakovsky, Jia Deng, Hao Su, Jonathan Krause, Sanjeev Satheesh, Sean Ma, Zhiheng Huang, Andrej Karpathy, Aditya Khosla, Michael Bernstein, Alexander~C. Berg, and Li~Fei-Fei. 2014.
\newblock \href {https://doi.org/10.48550/arXiv.1409.0575} {Imagenet large scale visual recognition challenge}.

\bibitem[{SenseTime(2024)}]{SenseNova}
SenseTime. 2024.
\newblock Sensenova.
\newblock \url{https://platform.sensenova.cn/home}.

\bibitem[{StepFun(2025)}]{stepfun}
StepFun. 2025.
\newblock Step-1o.
\newblock \url{https://platform.stepfun.com/}.

\bibitem[{Team et~al.(2023)Team, Anil, Borgeaud, Alayrac, Yu, Soricut, Schalkwyk, Dai, Hauth, Millican et~al.}]{team2023gemini}
Gemini Team, Rohan Anil, Sebastian Borgeaud, Jean-Baptiste Alayrac, Jiahui Yu, Radu Soricut, Johan Schalkwyk, Andrew~M Dai, Anja Hauth, Katie Millican, et~al. 2023.
\newblock Gemini: a family of highly capable multimodal models.
\newblock \emph{arXiv preprint arXiv:2312.11805}.

\bibitem[{Team(2024)}]{qvq-72b-preview}
Qwen Team. 2024.
\newblock \href {https://qwenlm.github.io/blog/qvq-72b-preview/} {Qvq: To see the world with wisdom}.

\bibitem[{Team(2025)}]{qwen2.5-VL}
Qwen Team. 2025.
\newblock \href {https://qwenlm.github.io/blog/qwen2.5-vl/} {Qwen2.5-vl}.

\bibitem[{Wang et~al.(2024{\natexlab{a}})Wang, Fu, Huang, Li, Liu, Liu, Ma, Xu, Zhou, Zhang, Yan, Mo, Liu, Lu, Li, Xiao, Chang, Roth, Zhang, Poon, and Chen}]{wang2024muirbenchcomprehensivebenchmarkrobust}
Fei Wang, Xingyu Fu, James~Y. Huang, Zekun Li, Qin Liu, Xiaogeng Liu, Mingyu~Derek Ma, Nan Xu, Wenxuan Zhou, Kai Zhang, Tianyi~Lorena Yan, Wenjie~Jacky Mo, Hsiang-Hui Liu, Pan Lu, Chunyuan Li, Chaowei Xiao, Kai-Wei Chang, Dan Roth, Sheng Zhang, Hoifung Poon, and Muhao Chen. 2024{\natexlab{a}}.
\newblock \href {https://arxiv.org/abs/2406.09411} {Muirbench: A comprehensive benchmark for robust multi-image understanding}.
\newblock \emph{Preprint}, arXiv:2406.09411.

\bibitem[{Wang et~al.(2024{\natexlab{b}})Wang, Fu, Huang, Li, Liu, Liu, Ma, Xu, Zhou, Zhang et~al.}]{wang2024muirbench}
Fei Wang, Xingyu Fu, James~Y Huang, Zekun Li, Qin Liu, Xiaogeng Liu, Mingyu~Derek Ma, Nan Xu, Wenxuan Zhou, Kai Zhang, et~al. 2024{\natexlab{b}}.
\newblock Muirbench: A comprehensive benchmark for robust multi-image understanding.
\newblock \emph{arXiv preprint arXiv:2406.09411}.

\bibitem[{Wang et~al.(2024{\natexlab{c}})Wang, Pan, Shi, Lu, Zhan, and Li}]{wang2024measuring}
Ke~Wang, Junting Pan, Weikang Shi, Zimu Lu, Mingjie Zhan, and Hongsheng Li. 2024{\natexlab{c}}.
\newblock \href {https://arxiv.org/abs/2402.14804} {Measuring multimodal mathematical reasoning with math-vision dataset}.
\newblock \emph{Preprint}, arXiv:2402.14804.

\bibitem[{Wang et~al.(2024{\natexlab{d}})Wang, Bai, Tan, Wang, Fan, Bai, Chen, Liu, Wang, Ge, Fan, Dang, Du, Ren, Men, Liu, Zhou, Zhou, and Lin}]{Qwen2VL}
Peng Wang, Shuai Bai, Sinan Tan, Shijie Wang, Zhihao Fan, Jinze Bai, Keqin Chen, Xuejing Liu, Jialin Wang, Wenbin Ge, Yang Fan, Kai Dang, Mengfei Du, Xuancheng Ren, Rui Men, Dayiheng Liu, Chang Zhou, Jingren Zhou, and Junyang Lin. 2024{\natexlab{d}}.
\newblock Qwen2-vl: Enhancing vision-language model's perception of the world at any resolution.
\newblock \emph{arXiv preprint arXiv:2409.12191}.

\bibitem[{Wang et~al.(2024{\natexlab{e}})Wang, Bai, Tan, Wang, Fan, Bai, Chen, Liu, Wang, Ge et~al.}]{wang2024qwen2}
Peng Wang, Shuai Bai, Sinan Tan, Shijie Wang, Zhihao Fan, Jinze Bai, Keqin Chen, Xuejing Liu, Jialin Wang, Wenbin Ge, et~al. 2024{\natexlab{e}}.
\newblock Qwen2-vl: Enhancing vision-language model's perception of the world at any resolution.
\newblock \emph{arXiv preprint arXiv:2409.12191}.

\bibitem[{Wang et~al.(2024{\natexlab{f}})Wang, Chen, Wang, Cao, Liu, Gao, Zhu, Zhu, Lu, Qiao, and Dai}]{wang2024mpo}
Weiyun Wang, Zhe Chen, Wenhai Wang, Yue Cao, Yangzhou Liu, Zhangwei Gao, Jinguo Zhu, Xizhou Zhu, Lewei Lu, Yu~Qiao, and Jifeng Dai. 2024{\natexlab{f}}.
\newblock Enhancing the reasoning ability of multimodal large language models via mixed preference optimization.
\newblock \emph{arXiv preprint arXiv:2411.10442}.

\bibitem[{White et~al.(2024)White, Dooley, Roberts, Pal, Feuer, Jain, Shwartz-Ziv, Jain, Saifullah, Naidu, Hegde, LeCun, Goldstein, Neiswanger, and Goldblum}]{livebench}
Colin White, Samuel Dooley, Manley Roberts, Arka Pal, Ben Feuer, Siddhartha Jain, Ravid Shwartz-Ziv, Neel Jain, Khalid Saifullah, Siddartha Naidu, Chinmay Hegde, Yann LeCun, Tom Goldstein, Willie Neiswanger, and Micah Goldblum. 2024.
\newblock \href {arXiv preprint arXiv:2406.19314} {Livebench: A challenging, contamination-free llm benchmark}.

\bibitem[{Wu et~al.(2024)Wu, Chen, Pan, Liu, Liu, Dai, Gao, Ma, Wu, Wang, Xie, Wu, Hu, Wang, Sun, Li, Piao, Guan, Liu, Xie, You, Dong, Yu, Zhang, Zhao, Wang, and Ruan}]{wu2024deepseekvl2mixtureofexpertsvisionlanguagemodels}
Zhiyu Wu, Xiaokang Chen, Zizheng Pan, Xingchao Liu, Wen Liu, Damai Dai, Huazuo Gao, Yiyang Ma, Chengyue Wu, Bingxuan Wang, Zhenda Xie, Yu~Wu, Kai Hu, Jiawei Wang, Yaofeng Sun, Yukun Li, Yishi Piao, Kang Guan, Aixin Liu, Xin Xie, Yuxiang You, Kai Dong, Xingkai Yu, Haowei Zhang, Liang Zhao, Yisong Wang, and Chong Ruan. 2024.
\newblock \href {https://arxiv.org/abs/2412.10302} {Deepseek-vl2: Mixture-of-experts vision-language models for advanced multimodal understanding}.
\newblock \emph{Preprint}, arXiv:2412.10302.

\bibitem[{Yue et~al.(2024)Yue, Ni, Zhang, Zheng, Liu, Zhang, Stevens, Jiang, Ren, Sun, Wei, Yu, Yuan, Sun, Yin, Zheng, Yang, Liu, Huang, Sun, Su, and Chen}]{yue2023mmmu}
Xiang Yue, Yuansheng Ni, Kai Zhang, Tianyu Zheng, Ruoqi Liu, Ge~Zhang, Samuel Stevens, Dongfu Jiang, Weiming Ren, Yuxuan Sun, Cong Wei, Botao Yu, Ruibin Yuan, Renliang Sun, Ming Yin, Boyuan Zheng, Zhenzhu Yang, Yibo Liu, Wenhao Huang, Huan Sun, Yu~Su, and Wenhu Chen. 2024.
\newblock Mmmu: A massive multi-discipline multimodal understanding and reasoning benchmark for expert agi.
\newblock In \emph{Proceedings of CVPR}.

\bibitem[{Zhang et~al.(2024{\natexlab{a}})Zhang, Da, Lee, Robinson, Wu, Song, Zhao, Raja, Slack, Lyu et~al.}]{zhang2024careful}
Hugh Zhang, Jeff Da, Dean Lee, Vaughn Robinson, Catherine Wu, Will Song, Tiffany Zhao, Pranav Raja, Dylan Slack, Qin Lyu, et~al. 2024{\natexlab{a}}.
\newblock A careful examination of large language model performance on grade school arithmetic.
\newblock \emph{arXiv preprint arXiv:2405.00332}.

\bibitem[{Zhang et~al.(2024{\natexlab{b}})Zhang, Jiang, Zhang, Lin, Guo, Qiu, Zhou, Lu, Chang, Qiao et~al.}]{zhang2024mathverse}
Renrui Zhang, Dongzhi Jiang, Yichi Zhang, Haokun Lin, Ziyu Guo, Pengshuo Qiu, Aojun Zhou, Pan Lu, Kai-Wei Chang, Yu~Qiao, et~al. 2024{\natexlab{b}}.
\newblock Mathverse: Does your multi-modal llm truly see the diagrams in visual math problems?
\newblock In \emph{European Conference on Computer Vision}, pages 169--186. Springer.

\bibitem[{Zhu et~al.(2023)Zhu, Chen, Shen, Li, and Elhoseiny}]{zhu2023minigpt}
Deyao Zhu, Jun Chen, Xiaoqian Shen, Xiang Li, and Mohamed Elhoseiny. 2023.
\newblock Minigpt-4: Enhancing vision-language understanding with advanced large language models.
\newblock \emph{arXiv preprint arXiv:2304.10592}.

\end{thebibliography}

\appendix

\section{Human Experiments}
\label{sec:human_experiments}
\subsection{Participant Recruitment}
Participants were primarily recruited by widely sharing the survey link on social media platforms, targeting a diverse audience. Additionally, personal networks were used to broaden the participant pool. Furthermore, offline testing was conducted, with participants completing surveys under researcher supervision. Notably, in all tests, we ensured that minors participated under the supervision of guardians. Each participant received a compensation of 2 USD, which is deemed fair and reasonable considering the estimated completion time of 20 minutes per survey.

\subsection{Participants Statistics}

The education level distribution of the participants is shown in Table~\ref{tab:degree_distribution}. The age and gender distribution of participants is shown in Figure~\ref{fig:age_gender_distribution}. 
\label{sec:human_experiments_statistics}

\begin{table}
  \centering
  \begin{tabular}{l c c}
    \toprule
    \textbf{Education} & \textbf{Number} & \textbf{Percentage} \\
    \midrule
    Bachelor & 233 & 40.5\% \\
    Associate & 106 & 18.4\% \\
    High School & 74 & 12.9\% \\
    Underage & 61 & 10.6\% \\
    Master or above & 58 & 10.1\% \\
    Junior High School & 35 & 6.1\% \\
    Primary School & 8 & 1.4\% \\
    \midrule
    Total & 575 &  \\
    \bottomrule
  \end{tabular}
  \caption{Education background of the participants.}
  \label{tab:degree_distribution}
\end{table}

\subsection{Experiments Interface}
\label{sec:human_experiments_interface}

\begin{figure}[h]
    \centering
    \includegraphics[width=\columnwidth]{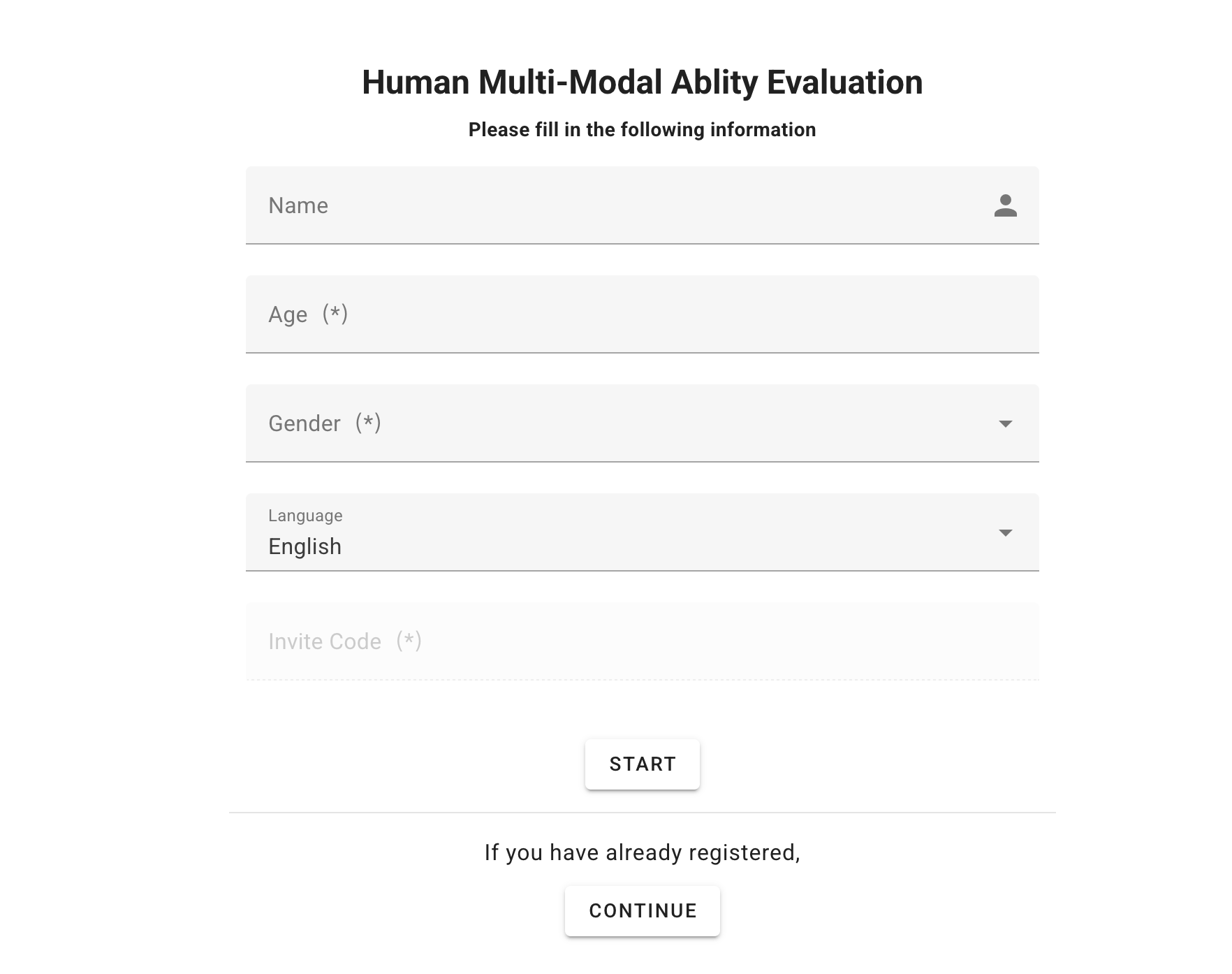}
    \caption{Screenshot of the User Personal Information Form: Capture essential details like name, age and gender.}
    \label{fig:user_interface} 
  \end{figure}
  \begin{figure}[h]
    \centering
    \includegraphics[width=0.8\columnwidth]{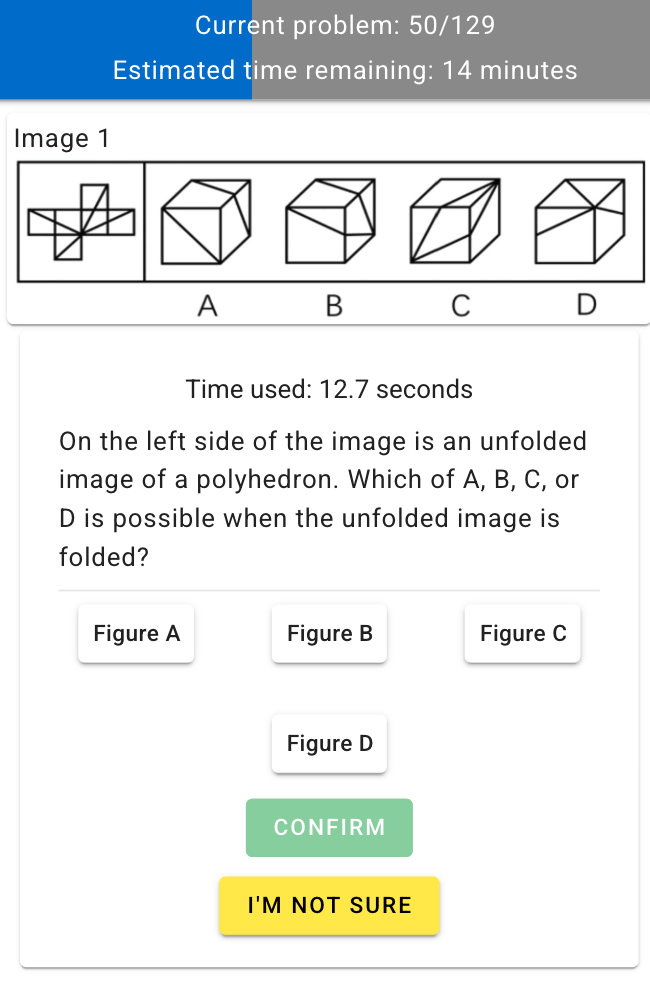}
    \caption{Screenshot of the User Question Interface: Engage with and respond to a series of questions in an organized format.}
    \label{fig:user_interface_question}
  \end{figure}

Users can access the human experiments through a web interface. The interface is divided into two sections: the User Personal Information Form (Figure~\ref{fig:user_interface}) and the User Question Interface (Figure~\ref{fig:user_interface_question}). The User Personal Information Form captures essential details like name, age, gender, and education level. The User Question Interface allows users to engage with and respond to a series of questions in a structured format.  

\subsection{Experiment Procedure}
Before beginning the experiments, users are required to complete the User Personal Information Form. Upon submitting the form, users are provided with the following three instructions: 1) Do not search for any information during the test. 
2) Avoid distractions during the test. If necessary, you may temporarily exit and re-enter the test interface later.
3) If the image is unclear, click to enlarge it.

The questions are displayed in a random order, and each question is accompanied by a timer. While there is no time limit for completing the test, users may choose to skip questions. However, once a question is answered or skipped, it cannot be revisited.

\subsection{Information of the Best Human Performance}
\label{sec:best_human_performance}
The best human performance sample was selected from the first stage of the human experiments. The individual is a 21-year-old Chinese student, with a strong background in mathematics and computer science. With prior experience in algorithm competitions, they demonstrates exceptional reasoning abilities and the capacity to solve complex problems both quickly and accurately. Additionally, they achieved the highest GPA in their class. It is important to note that this sample does not represent the highest level of expertise globally but reflects the best human performance observed within the scope of our experiments.

The best human performance in each category is shown in Table~\ref{tab:best_human_performance}. They was tested on the Chinese subset of the MANBench dataset. They achieved a perfect score in the Image Consistent category. They answered a total of 1021 questions, which is 77.7\% of the total questions in MANBench. Their total time spent on the test is 14.3 hours. Their performance is significantly higher than the average human performance across all tasks.
\begin{table}[t]
    \small
    \renewcommand{\arraystretch}{0.7} 
\centering
\begin{tabular}{lcc}
  \toprule
\textbf{\footnotesize{Category}} & \makecell{\textbf{\footnotesize{Correct/Total}}} & \makecell{\textbf{\footnotesize{Score}}}\\
  \midrule
\footnotesize{Text-image Understanding} & \footnotesize{\makecell{89/100}} & \footnotesize{\makecell{89.00}} \\
\footnotesize{Image Consistency} & \footnotesize{\makecell{158/158}} & \footnotesize{\makecell{100.00}} \\
\footnotesize{Numbers Pattern} & \footnotesize{\makecell{123/134}} & \footnotesize{\makecell{91.79}} \\
\footnotesize{Spatial Imagination} & \footnotesize{\makecell{174/184}} & \footnotesize{\makecell{94.57}} \\
\footnotesize{Text Locating} & \footnotesize{\makecell{101/107}} & \footnotesize{\makecell{94.39}} \\
\footnotesize{Multi-image Understanding} & \footnotesize{\makecell{78/84}} & \footnotesize{\makecell{92.86}} \\
\footnotesize{Knowledge} & \footnotesize{\makecell{74/99}} & \footnotesize{\makecell{74.75}} \\
\footnotesize{Puzzles} & \footnotesize{\makecell{70/81}} & \footnotesize{\makecell{86.42}} \\
\footnotesize{Transmorphic Understanding} & \footnotesize{\makecell{67/74}} & \footnotesize{\makecell{90.54}} \\
\textbf{\footnotesize{Average}} & & \makecell{\footnotesize{90.48}} \\ 
  \bottomrule
\end{tabular}
\caption{Best human performance on each category.}
\label{tab:best_human_performance}
\end{table}
  
\section{MLLM Experiments}
\label{sec:mlm_experiments}
All evaluations of open-source models were conducted using four NVIDIA A800 GPUs. The execution time for each model varied significantly depending on its architectural complexity and number of parameters. In total, the comprehensive evaluation of all open-source MLLMs required approximately 200 A800-hours.

Closed-source models were evaluated using publicly available APIs. The version of each closed-source model is listed in Table~\ref{tab:model_version}.
\begin{table}
    \centering
    \begin{tabular}{l l}
      \toprule
      \textbf{Model Name} & \textbf{Version / Test Time} \\
      \midrule
      Claude-3.5-Sonnet & claude-3-5-sonnet-20241022 \\
      Gemini-1.5-Pro & gemini-1.5-pro-002 \\
      SenseNova & 2025-02-08 \\
      Step-1o & step-1o-vision-32k \\
      GPT-4o & gpt-4o-2024-11-20\\
      GPT-o1 & gpt-o1-2024-12-17\\
      \bottomrule
    \end{tabular}
    \caption{Closed-source model names and their corresponding version or test time.}
    \label{tab:model_version}
  \end{table}
  
\section{Analysis}
\subsection{Chinese Subset Results}
\label{sec:chinese_subset_results}
\begin{figure}[h]
    \includegraphics[width=\columnwidth]{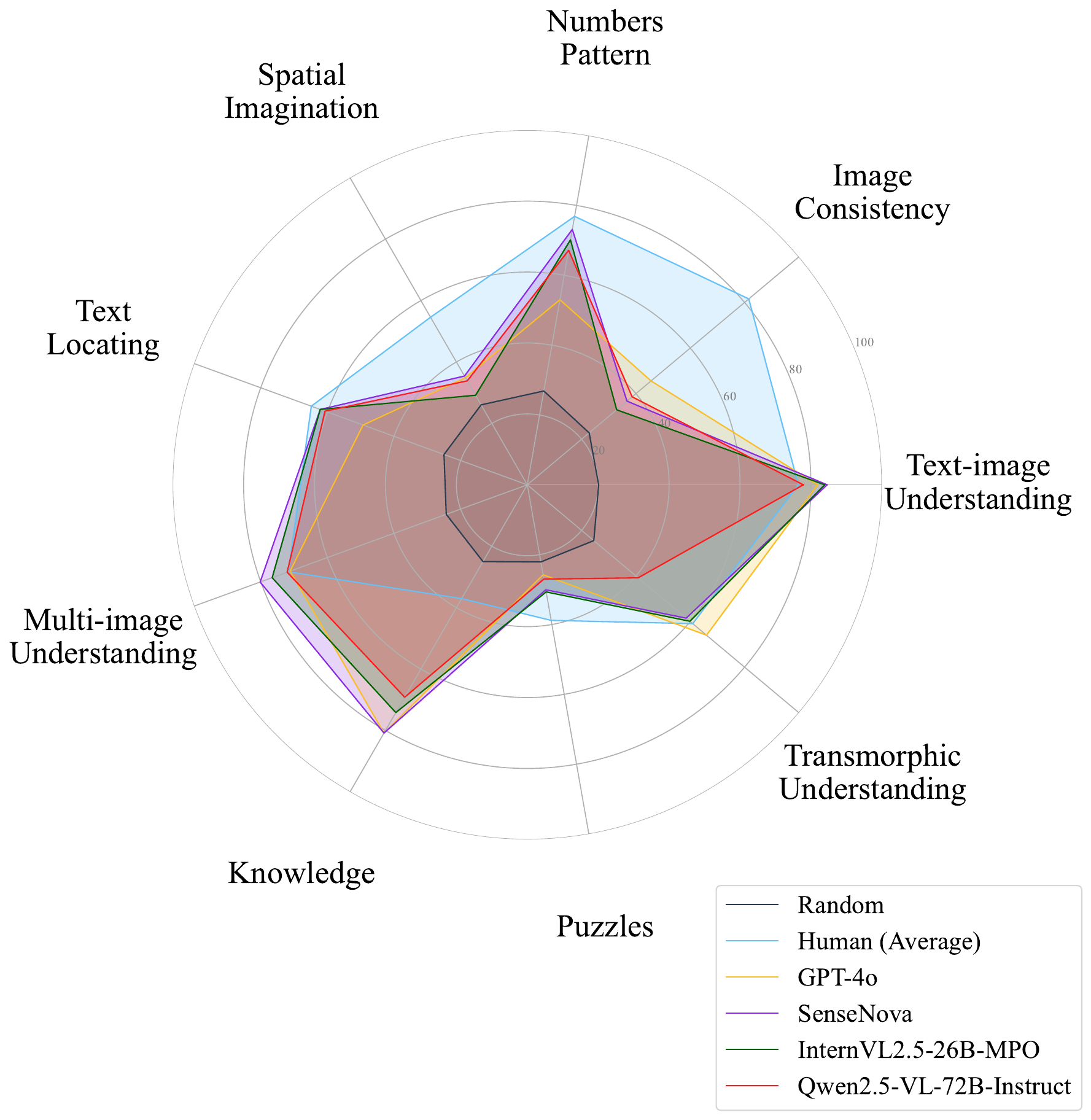}
    \caption{Performance comparison on the MANBench Chinese subset among human average accuracy, and some selected MLLMs. Please refer to Table~\ref{tab:performance_chinese} for more results.}
    \label{fig:radar_cn} 
  \end{figure}

The results of MLLMs on the Chinese subset are shown in Table~\ref{tab:performance_chinese} and Figure~\ref{fig:radar_cn}.
\begin{table*}
  \centering
\resizebox{\textwidth}{!}{
  \begin{tabular}{lcccccccccc}
    \toprule[1.2pt]
   & \textbf{\scriptsize{\makecell{Overall}}} & \textbf{\scriptsize{\makecell{Text-image \\ Understanding \\ (149)}}} & \textbf{\scriptsize{\makecell{Image \\ Consistency \\ (158)}}} & \textbf{\scriptsize{\makecell{Numbers \\ Pattern \\ (134)}}} & \textbf{\scriptsize{\makecell{Spatial \\ Imagination \\ (192)}}} & \textbf{\scriptsize{\makecell{Text \\ Locating \\ (140)}}} & \textbf{\scriptsize{\makecell{Multi-image \\ Understanding \\ (111)}}} & \textbf{\scriptsize{\makecell{Knowledge \\ (120)}}} & \textbf{\scriptsize{\makecell{Puzzles \\ (163)}}} & \textbf{\scriptsize{\makecell{Transmorphic \\ Understanding \\ (147)}}} \\
   \midrule[1.2pt]
    Human (Average) & 62.26 & 76.46 & 81.55 & 76.88 & 54.56 & 64.82 & 71.69 & 37.05 & 38.83 & 60.90 \\
    Human (Best) & 90.87 & 89.00 & 100.00 & 91.79 & 94.57 & 94.39 & 92.86 & 74.75 & 86.42 & 90.54 \\
    Random & 24.05 & 20.13 & 22.78 & 26.87 & 26.04 & 25.00 & 24.32 & 25.00 & 22.09 & 24.49 \\
\hline \multicolumn{11}{c}{ \textbf{Open-source  MLLMs}} \\ \hline 
    Deepseek-VL2 & 43.15 & 73.83 & 26.58 & 57.46 & 29.69 & 42.86 & 38.74 & \underline{66.67} & 22.09 & 42.18 \\
    InternVL2-26B & 40.18 & 61.07 & 29.11 & 55.22 & 32.29 & 31.43 & 34.23 & \underline{57.50} & 25.77 & 42.18 \\
    InternVL2-8B & 34.47 & 50.34 & 24.68 & 41.79 & 26.04 & 26.43 & 45.05 & \underline{49.17} & 22.09 & 34.69 \\
    InternVL2.5-26B-MPO & 55.25 & \underline{83.89} & 32.91 & 70.15 & 29.17 & 62.14 & \underline{76.58} & \underline{74.17} & 30.67 & 59.86 \\
    InternVL2.5-78B-MPO & \textbf{60.12} & \textbf{\underline{88.59}} & 37.97 & \textbf{75.37} & \textbf{37.50} & \textbf{\underline{65.71}} & \textbf{\underline{78.38}} & \textbf{\underline{83.33}} & 33.74 & \textbf{\underline{61.90}} \\
    QVQ-72B-Preview & 49.85 & 73.83 & 25.32 & 61.19 & 32.29 & 48.57 & 64.86 & \underline{74.17} & \textbf{36.81} & 48.98 \\
    Qwen2-VL-72B-Instruct & 48.33 & \underline{78.52} & 25.95 & 63.43 & 31.25 & 45.00 & 57.66 & \underline{73.33} & 25.15 & 51.70 \\
    Qwen2.5-VL-72B-Instruct & 52.05 & \underline{77.85} & \textbf{38.61} & 67.16 & 33.85 & 60.71 & \underline{72.07} & \underline{69.17} & 26.99 & 40.82 \\
\hline \multicolumn{11}{c}{ \textbf{Closed-source MLLMs}} \\ \hline 
    Claude-3.5-Sonnet & 54.79 & \underline{83.22} & 44.30 & 64.18 & \textbf{36.98} & 43.57 & 69.37 & \underline{75.00} & 26.99 & \underline{65.99} \\
    GPT-4o & 54.49 & \underline{81.88} & 45.57 & 52.99 & 34.90 & 49.29 & 71.17 & \textbf{\underline{80.83}} & 25.77 & \underline{65.99} \\
    GPT-o1 & \textbf{60.12} & \textbf{\underline{88.59}} & \textbf{56.96} & 72.39 & 33.33 & 53.57 & 70.27 & \textbf{\underline{80.83}} & 30.67 & \textbf{\underline{72.79}} \\
    Gemini-1.5-Pro & 52.66 & \underline{81.88} & 39.24 & 55.22 & 29.69 & 50.71 & 67.57 & \underline{79.17} & \textbf{31.29} & 57.82 \\
    SenseNova & 57.69 & \underline{84.56} & 36.71 & \textbf{73.13} & 35.42 & \textbf{62.14} & \textbf{\underline{80.18}} & \textbf{\underline{80.83}} & 30.06 & 58.50 \\
    Step-1o & 54.79 & \underline{84.56} & 33.54 & 69.40 & 33.85 & 50.00 & 68.47 & \underline{76.67} & 28.83 & \underline{66.67} \\
     \bottomrule
  \end{tabular}
}
\caption{Results of different models on the MANBench Chinese subset. The first row shows task names and number of test data. The best performance in each task is in bold. Values exceeding human average performance are underlined. For the information of the best human performance, please refer to Appendix~\ref{sec:best_human_performance}. The overall score is calculated by weighting the task scores according to the number of questions in each task.}
\label{tab:performance_chinese}
\end{table*}

\subsection{Ablation Study}
\label{sec:ablation_study}
Since the MLLMs tend to refuse to answer the questions when the images are removed, we use a prompt to compel the MLLMs to select an answer. The prompt is \texttt{“Even if the images are not provided or you are not sure about the answer, you are forced to choose one of the options”}. The results are shown in Figure~\ref{fig:ablation}.

\section{Artifact Licensing and Distribution}
\label{sec:artifact_licensing}
\noindent\textbf{Data Licensing: }MANBench utilizes images from MUIRBench~\cite{wang2024muirbench}, ScienceQA~\citep{lu2022learn}, MMStar~\citep{chen2024we}, OxfordPets~\citep{oxfordpets}, MMBench~\citep{MMbench}, and BLINK~\citep{fu2024blink}. MUIRBench, ScienceQA, MMStar, and OxfordPets are licensed under the Creative Commons Attribution-NonCommercial-ShareAlike 4.0 International License (CC BY-NC-SA 4.0). MMBench and BLINK are licensed under the Apache License 2.0. MANBench is released under the CC BY-NC-SA 4.0 license, ensuring full compliance with the copyright requirements of the source datasets. All images included in this paper have been carefully reviewed to ensure compliance with applicable copyright laws and are accompanied by proper attribution. If you are the copyright holder of any image featured in this work and believe its usage violates your licensing terms, please contact us promptly. We are committed to addressing any legitimate copyright concerns in a timely and respectful manner.

\noindent\textbf{Code Licensing: }The evaluation code of MANBench is developed based on VLMEvalKit~\citep{duan2024vlmevalkit}, licensed under the Apache License 2.0. MANBench's code is also released under the Apache License 2.0, ensuring open-source accessibility for the community. The code and dataset are available at \href{https://anonymous.4open.science/r/MANBench-4742/}{https://anonymous.4open.science/r/MANBench-4742/}.
\section{Data Collection and Privacy Protection Measures}
Our dataset comprises images and questions that were meticulously curated and rigorously evaluated to ensure the exclusion of sensitive or inappropriate content. Prior to finalizing the dataset, we conducted a pilot study to assess the quality of the questions and images. This pilot involved a small cohort of participants who provided feedback on the clarity, relevance, and potential offensiveness of the materials. Their feedback was instrumental in refining the dataset to align with the highest standards of quality and ethical integrity.

Demographic information, including optional fields such as name, age, gender, and education level, was collected from participants. This information were securely stored in an encrypted database with access restricted to authorized personnel, ensuring full compliance with data protection regulations.

To uphold the ethical integrity of the research and protect participant privacy, rigorous anonymization protocols were implemented before the release of the experimental dataset. Specifically, all personally identifiable information (PII) was systematically removed, and data points were carefully processed to prevent any potential re-identification or traceability to individual participants. Furthermore, we strictly adhered to ethical research practices by obtaining written informed consent from all participants prior to data collection.
\clearpage

\section{Datasheet for MANBench}

\newcommand\squeezeitem{\vspace{-0mm}}

\datasheetsection{Motivation for Dataset Creation}

\begin{datasheetitem}{Why was the dataset created?}[(e.g., were there specific tasks in mind, or a specific gap that needed to be filled?)]
    The dataset, \textbf{MANBench}, was created to address significant limitations in existing benchmarks for evaluating multimodal capabilities of both humans and Multimodal Large Language Models (MLLMs). Existing benchmarks often focus on narrow tasks, require domain-specific knowledge, and rely on insufficient human evaluations characterized by small sample sizes and limited diversity. MANBench aims to provide a comprehensive and fair assessment framework to rigorously compare human and MLLM performance across a wide range of multimodal reasoning tasks.

\squeezeitem\end{datasheetitem}

\begin{datasheetitem}{What (other) tasks could the dataset be used for?}[Are there obvious tasks for which it should \emph{not} be used?]
MANBench could be used for tasks such as:
\begin{itemize}
    \item Multimodal reasoning: Evaluating the ability to integrate textual and visual information.
    \item Cross-modal understanding: Assessing tasks such as Transmorphic Understanding, Image Consistency, and Multi-image Understanding.
    \item Complex problem-solving: Testing capabilities in highly complex tasks, including Puzzles and Spatial Imagination.
\end{itemize}

MANBench is not suitable for tasks that rely extensively on domain-specific knowledge or for simple retrieval tasks, as it is specifically designed to emphasize reasoning and multimodal fusion.
\squeezeitem\end{datasheetitem}

\begin{datasheetitem}{Has the dataset been used for any tasks already?}[If so, where are the results so  others can compare (e.g., links to published papers)?]
    Yes, the MANBench dataset has already been utilized for tasks involving both Multimodal Large Language Models (MLLMs) and human participants. Specifically:

    \textbf{MLLM Evaluation:}
    The dataset was employed to evaluate 12 state-of-the-art MLLMs, including GPT-4o, Qwen2.5-VL-72B-Instruct, and Gemini-1.5 Pro. The results revealed that while MLLMs excel in tasks such as Knowledge and Text-Image Understanding, they struggle in tasks requiring deeper cross-modal reasoning, such as Transmorphic Understanding, Image Consistency, and Multi-image Understanding.
    
    \textbf{Human Testing:}
    The dataset was also used to conduct large-scale human evaluations. A total of 575 participants from diverse backgrounds were recruited to complete the tasks in MANBench. These human evaluations established a performance baseline for comparing MLLMs and highlighted areas where both humans and MLLMs encounter similar difficulties, such as Puzzles and Spatial Imagination.

\squeezeitem\end{datasheetitem}

\begin{datasheetitem}{Who funded the creation of the dataset?}[If there is an associated grant, provide the grant number.]
    No external funding was used in the creation of the MANBench dataset.
\squeezeitem\end{datasheetitem}

\datasheetsection{Dataset Composition}

\begin{datasheetitem}{What are the instances?}[(that is, examples; e.g., documents, images, people, countries) Are there multiple types of instances? (e.g., movies, users, ratings; people, interactions between them; nodes, edges)]
    The instances in MANBench consist of 1,314 bilingual(English and Chinese) questions and 2,231 images, organized across nine tasks which are in visual question answer(VQA) format. 
    
\squeezeitem\end{datasheetitem}

\begin{datasheetitem}{Are relationships between instances made explicit in the data}[(e.g., social network links, user/movie ratings, etc.)?]

	The relationships between instances are implicit rather than explicit. Each question and its associated image(s) are designed to evaluate specific multimodal reasoning abilities. However, there are no explicit relationships, such as social network links or user/movie ratings, between the instances.
\squeezeitem\end{datasheetitem}

\begin{datasheetitem}{How many instances of each type are there?}
    The MANBench benchmark comprises a total of 1,314 questions distributed across nine distinct multimodal tasks, each designed to rigorously evaluate the capabilities of both humans and Multimodal Large Language Models (MLLMs). The distribution of tasks is as follows: Text-image Understanding (149 instances), Image Consistency (158 instances), Numbers Pattern (134 instances), Spatial Imagination (192 instances), Text Locating (140 instances), Multi-image Understanding (111 instances), Knowledge (120 instances), Puzzles (163 instances), and Transmorphic Understanding (147 instances). 
    
\squeezeitem\end{datasheetitem}

\begin{datasheetitem}{What data does each instance consist of?}[``Raw'' data (e.g., unprocessed text or images)? Features/attributes? Is there a label/target associated with instances? If the instances are related to people, are subpopulations identified (e.g., by age, gender, etc.) and what is their distribution?]
	Each instance consists of textual questions (including the prompt, options, answers, and the task to which it belongs) and images. 
\squeezeitem\end{datasheetitem}

\begin{datasheetitem}{Is everything included or does the data rely on external resources?}[(e.g., websites, tweets, datasets) If external resources, a) are there guarantees that they will exist, and remain constant, over time; b) is there an official archival version. Are there licenses, fees or rights associated with \emph{any} of the data?] 
    Everything included in the MANBench dataset is self-contained and does not rely on external resources.
\squeezeitem\end{datasheetitem}

\begin{datasheetitem}{Are there recommended data splits or evaluation measures?}[(e.g., training, development, testing; accuracy/AUC)] 
    We follow standard setups as in the
VLMEvalKit~\citep{duan2024vlmevalkit}, where the temperature is set to 0 and retry is set to 10.
\squeezeitem\end{datasheetitem}

\begin{datasheetitem}{What experiments were initially run on this dataset?}[Have a summary of those results and, if available, provide the link to a paper with more information here.]
    The initial experiments involved evaluating 12 state-of-the-art MLLMs on MANBench and comparing their performance to that of 575 human participants across 9 tasks. The results showed that MLLMs excel in tasks like Knowledge and Text-Image Understanding but struggle in tasks requiring deeper cross-modal reasoning.
\squeezeitem\end{datasheetitem}

\datasheetsection{Data Collection Process}

\begin{datasheetitem}{How was the data collected?}[(e.g., hardware apparatus/sensor, manual human curation, software program, software interface/API; how were these constructs/measures/methods validated?)]
    The MANBench dataset was collected by authors manually.
\squeezeitem\end{datasheetitem}

\begin{datasheetitem}{Who was involved in the data collection process?}[(e.g., students, crowdworkers) How were they compensated? (e.g., how much were crowdworkers paid?)]
    Only the authors were involved in the data collection process. No crowdworkers or external contributors were used.
\squeezeitem\end{datasheetitem}

\begin{datasheetitem}{Over what time-frame was the data collected?}[Does the collection time-frame match the creation time-frame?]
    The MANBench dataset was collected over a period of 3 months. The collection time-frame matches the creation time-frame.
\squeezeitem\end{datasheetitem}

\begin{datasheetitem}{How was the data associated with each instance acquired?}[Was the data directly observable (e.g., raw text, movie ratings), reported by subjects (e.g., survey responses), or indirectly inferred/derived from other data (e.g., part of speech tags; model-based guesses for age or language)? If the latter two, were they validated/verified and if so how?]
   All the data associated with each instance was directly observable and manually curated by the authors. 
\squeezeitem\end{datasheetitem}

\begin{datasheetitem}{Does the dataset contain all possible instances?}[Or is it, for instance, a sample (not necessarily random) from a larger set of instances?]
    The MANBench dataset comprises a comprehensive set of instances designed to evaluate a wide range of multimodal reasoning tasks. However, it is not exhaustive and may not cover all possible multimodal reasoning scenarios.
\squeezeitem\end{datasheetitem}

\begin{datasheetitem}{If the dataset is a sample, then what is the population?}[What was the sampling strategy (e.g., deterministic, probabilistic with specific sampling probabilities)? Is the sample representative of the larger set (e.g., geographic coverage)?  If not, why not (e.g., to cover a more diverse range of instances)?  How does this affect possible uses?]
    The MANBench dataset is not a sample but a comprehensive collection of instances designed to evaluate a wide range of multimodal reasoning tasks. The instances are carefully selected to cover a diverse range of multimodal reasoning scenarios and are representative of the larger set of multimodal reasoning tasks.
\squeezeitem\end{datasheetitem}

\begin{datasheetitem}{Is there information missing from the dataset and why?}[(this does not include intentionally dropped instances; it might include, e.g., redacted text, withheld documents) Is this data missing because it was unavailable?]
    No, there is no information missing from the dataset. All instances are complete and self-contained.
\squeezeitem\end{datasheetitem}

\begin{datasheetitem}{Are there any known errors, sources of noise, or redundancies in the data?}
    No, the MANBench dataset has been carefully curated to ensure accuracy and consistency. There are no known errors, sources of noise, or redundancies in the data. If any errors are identified, they will be corrected in future updates.
\squeezeitem\end{datasheetitem}

\datasheetsection{Data Preprocessing}

\begin{datasheetitem}{What preprocessing/cleaning was done?}[(e.g., discretization or bucketing, tokenization, part-of-speech tagging, SIFT feature extraction, removal of instances, processing of missing values, etc.)]
   We conducted manual preprocessing to ensure the quality of the data. 
\squeezeitem\end{datasheetitem}

\begin{datasheetitem}{Was the ``raw'' data saved in addition to the preprocessed/cleaned data?}[(e.g., to support unanticipated future uses)]
    Yes, the raw data was saved by the authors in addition to the preprocessed/cleaned data.
\squeezeitem\end{datasheetitem}

\begin{datasheetitem}{Is the preprocessing software available?}
    No, the preprocessing software is not available.
\squeezeitem\end{datasheetitem}

\begin{datasheetitem}{Does this dataset collection/processing procedure achieve the motivation for creating the dataset stated in the first section of this datasheet?}
    Yes, the dataset collection and processing procedures fulfill the motivation for creating the dataset by providing a comprehensive and fair assessment framework to rigorously compare human and Multimodal Large Language Model (MLLM) performance across a wide range of multimodal reasoning tasks.
\squeezeitem\end{datasheetitem}

\datasheetsection{Dataset Distribution}

\begin{datasheetitem}{How is the dataset distributed?}[(e.g., website, API, etc.; does the data have a DOI; is it archived redundantly?)]
    The MANBench dataset is distributed through the official \href{https://huggingface.co/datasets/MANBench/MANBench}{Hugging Face repository}.
\squeezeitem\end{datasheetitem}

\begin{datasheetitem}{When will the dataset be released/first distributed?}[(Is there a canonical paper/reference for this dataset?)]
The dataset will be released on Feb 10th, 2025.
\squeezeitem\end{datasheetitem}

\begin{datasheetitem}{What license (if any) is it distributed under?}[Are there any copyrights on the data?]
    The MANBench dataset is released under the Creative Commons Attribution-NonCommercial-ShareAlike 4.0 International License (CC BY-NC-SA 4.0).
\squeezeitem\end{datasheetitem}

\begin{datasheetitem}{Are there any fees or access/export restrictions?}
    No. The MANBench dataset is freely available for non-commercial use, sharing, and adaptation.
\squeezeitem\end{datasheetitem}

\datasheetsection{Dataset Maintenance}

\begin{datasheetitem}{Who is supporting/hosting/maintaining the dataset?}[How does one contact the owner/curator/manager of the dataset (e.g. email address, or other contact info)?]
    The first author of the paper “MANBench: Is Your Multimodal Model Smarter than Human?” is responsible for supporting, hosting, and maintaining the MANBench dataset. However, during the review process, due to the double-blind policy, we are unable to provide contact information. Once the paper is accepted, the contact information will be updated.
\squeezeitem\end{datasheetitem}

\begin{datasheetitem}{Will the dataset be updated?}[How often and by whom? How will updates/revisions be documented and communicated (e.g., mailing list, GitHub)? Is there an erratum?]
    Yes. The dataset may be updated periodically to address errors, improve the quality of the data, or add new instances. Updates will be documented and communicated through the official \href{https://huggingface.co/datasets/MANBench/MANBench}{Hugging Face repository}.
\squeezeitem\end{datasheetitem}

\begin{datasheetitem}{If the dataset becomes obsolete how will this be communicated?}
    We will communicate the obsolescence of the dataset through the official  \href{https://anonymous.4open.science/r/MANBench-4742/}{GitHub repository}.
\squeezeitem\end{datasheetitem}

\begin{datasheetitem}{Is there a repository to link to any/all papers/systems that use this dataset?}
    Yes, once the paper is accepted, we will provide a link to the official \href{https://anonymous.4open.science/r/MANBench-4742/}{GitHub repository} where all papers and systems utilizing the MANBench dataset will be linked.
\squeezeitem\end{datasheetitem}

\begin{datasheetitem}{If others want to extend/augment/build on this dataset, is there a mechanism for them to do so?}[If so, is there a process for tracking/assessing the quality of those contributions. What is the process for communicating/distributing these contributions to users?]
    Yes, we encourage researchers to extend, augment, and build on the MANBench dataset. MANBench is released under the CC BY-NC-SA 4.0 license, which permits non-commercial use, sharing, and adaptation. Researchers can contribute to the dataset by submitting pull requests to the official \href{https://huggingface.co/datasets/MANBench/MANBench}{Hugging Face repository}. All contributions will be reviewed by the dataset maintainers to ensure quality and relevance before being incorporated into the dataset.
\squeezeitem\end{datasheetitem}

\datasheetsection{Legal \& Ethical Considerations}

\begin{datasheetitem}{If the dataset relates to people (e.g., their attributes) or was generated by people, were they informed about the data collection?}[(e.g., datasets that collect writing, photos, interactions, transactions, etc.)] 
The MANBench dataset does not generate any data that relates to people, but we collected data related to people from existing datasets. 
\squeezeitem\end{datasheetitem}

\begin{datasheetitem}{If it relates to other ethically protected subjects, have appropriate obligations been met?} [(e.g., medical data might include information collected from animals)]
    No, the dataset does not relate to other ethically protected subjects.
\squeezeitem\end{datasheetitem}

\begin{datasheetitem}{If it relates to people, were there any ethical review applications/reviews/approvals?} [(e.g. Institutional Review Board applications)]
No, the dataset only conatins data related to people from existing datasets, which have already been reviewed and approved by the respective institutions.
\squeezeitem\end{datasheetitem}

\begin{datasheetitem}{If it relates to people, could this dataset expose people to harm or legal action?}[(e.g., financial social or otherwise) What was done to mitigate or reduce the potential for harm?]
    No, the dataset does not contain any information that could expose people to harm or legal action.
\squeezeitem\end{datasheetitem}

\begin{datasheetitem}{If it relates to people, does it unfairly advantage or disadvantage a particular social group?}[In what ways? How was this mitigated?]
    No, the dataset does not unfairly advantage or disadvantage any particular social group.
\squeezeitem\end{datasheetitem}

\begin{datasheetitem}{If it relates to people, were they provided with privacy guarantees?}[If so, what guarantees and how are these ensured?]
    Yes, all personally identifiable information (PII) was systematically removed, and data points were carefully processed to prevent any potential re-identification or traceability to individual participants. The dataset is designed to be fully anonymized and does not contain any sensitive information.
\squeezeitem\end{datasheetitem}

\begin{datasheetitem}{Does the dataset comply with the EU General Data Protection Regulation (GDPR)?}[Does it comply with any other standards, such as the US Equal Employment Opportunity Act?]
    Yes, the dataset complies with the EU General Data Protection Regulation (GDPR) and other relevant data protection standards.
\squeezeitem\end{datasheetitem}

\begin{datasheetitem}{Does the dataset contain information that might be considered sensitive or confidential?}[(e.g., personally identifying information)]
    No, the dataset does not contain any sensitive or confidential information.
\squeezeitem\end{datasheetitem}

\begin{datasheetitem}{Does the dataset contain information that might be considered inappropriate or offensive?}
    No, the dataset does not contain any inappropriate or offensive information.
\squeezeitem\end{datasheetitem}

\clearpage
\newpage
\begin{figure*}[h]
    \centering
    \includegraphics[width=\textwidth]{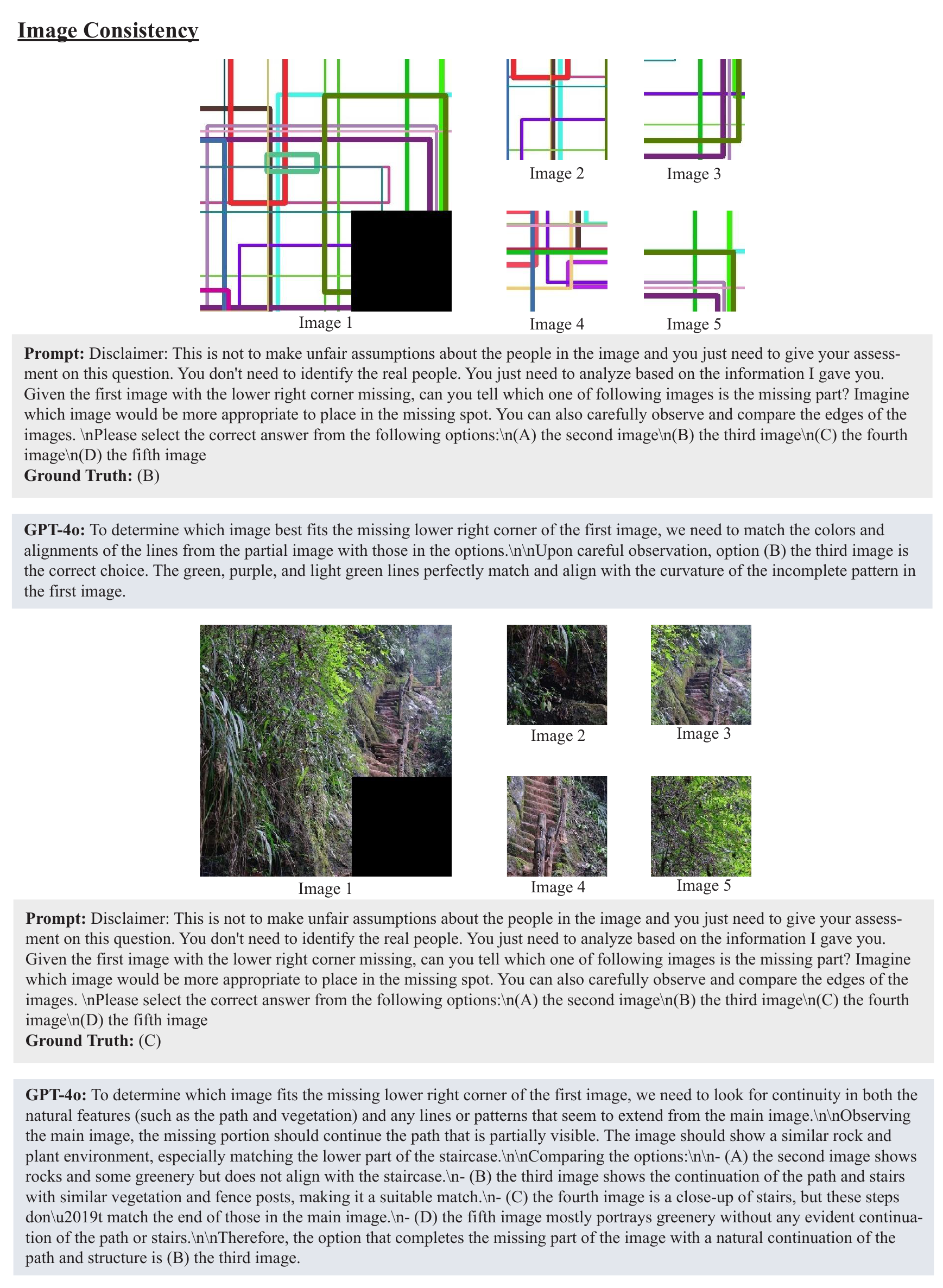}
    \caption{Examples of data in MANBench with GPT-4o predictions.(1/9)}
\end{figure*}
\newpage
\begin{figure*}[h]
    \centering
    \includegraphics[width=\textwidth]{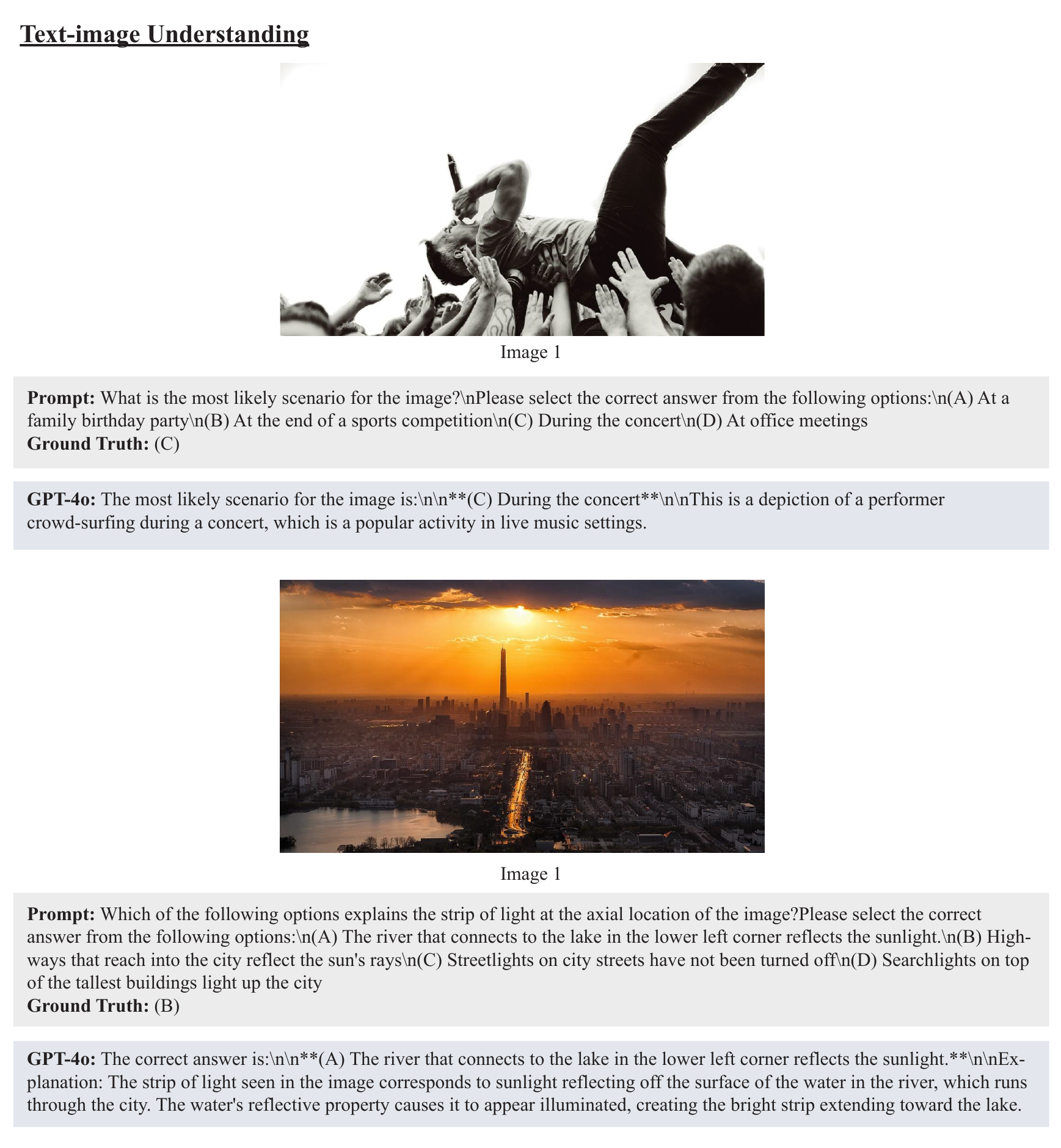}
    \caption{Examples of data in MANBench with GPT-4o predictions.(2/9)}
\end{figure*}
\newpage
\begin{figure*}[h]
    \centering
    \includegraphics[width=\textwidth]{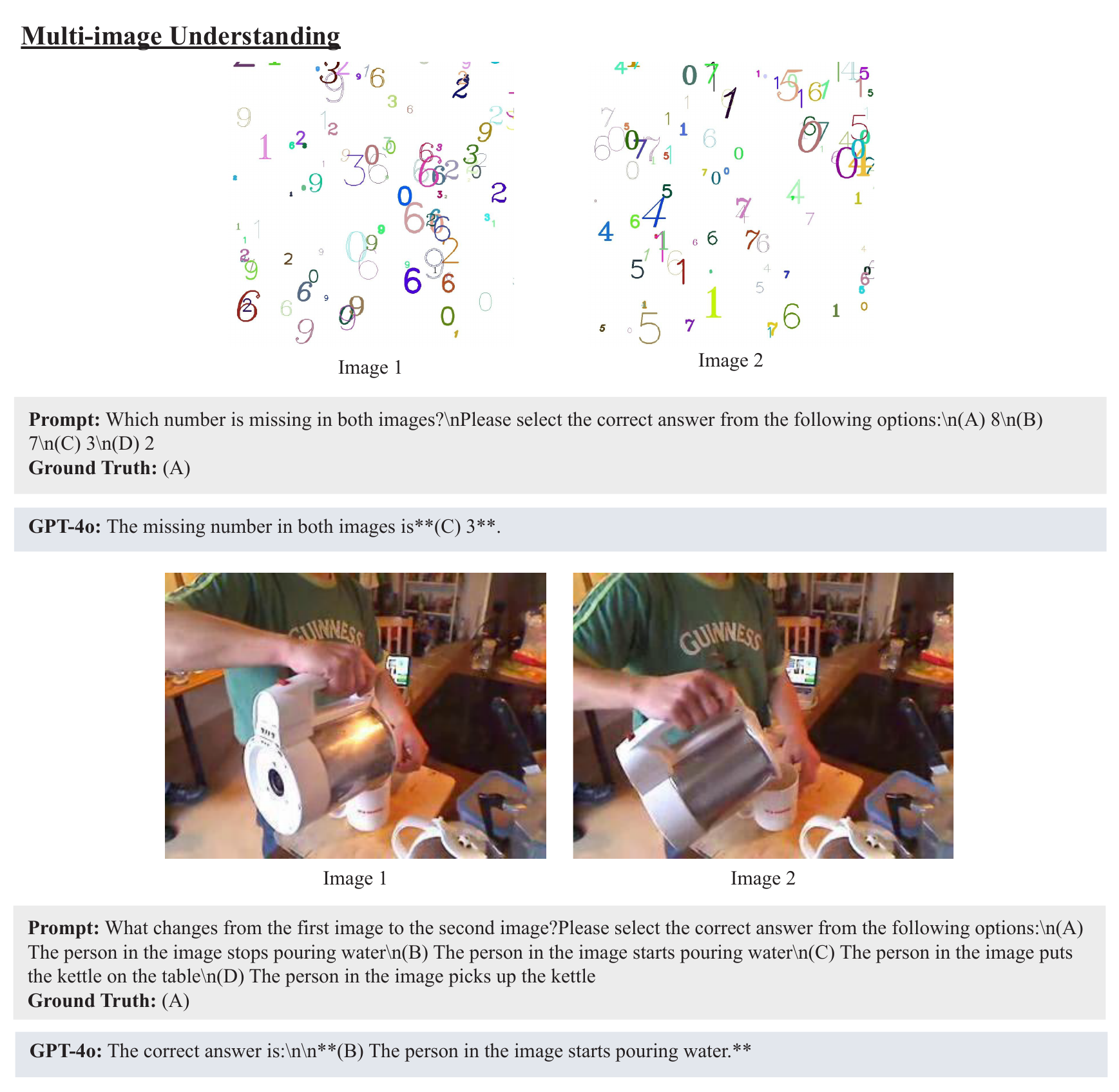}
    \caption{Examples of data in MANBench with GPT-4o predictions.(3/9)}
\end{figure*}
\newpage
\begin{figure*}[h]
    \centering
    \includegraphics[width=\textwidth]{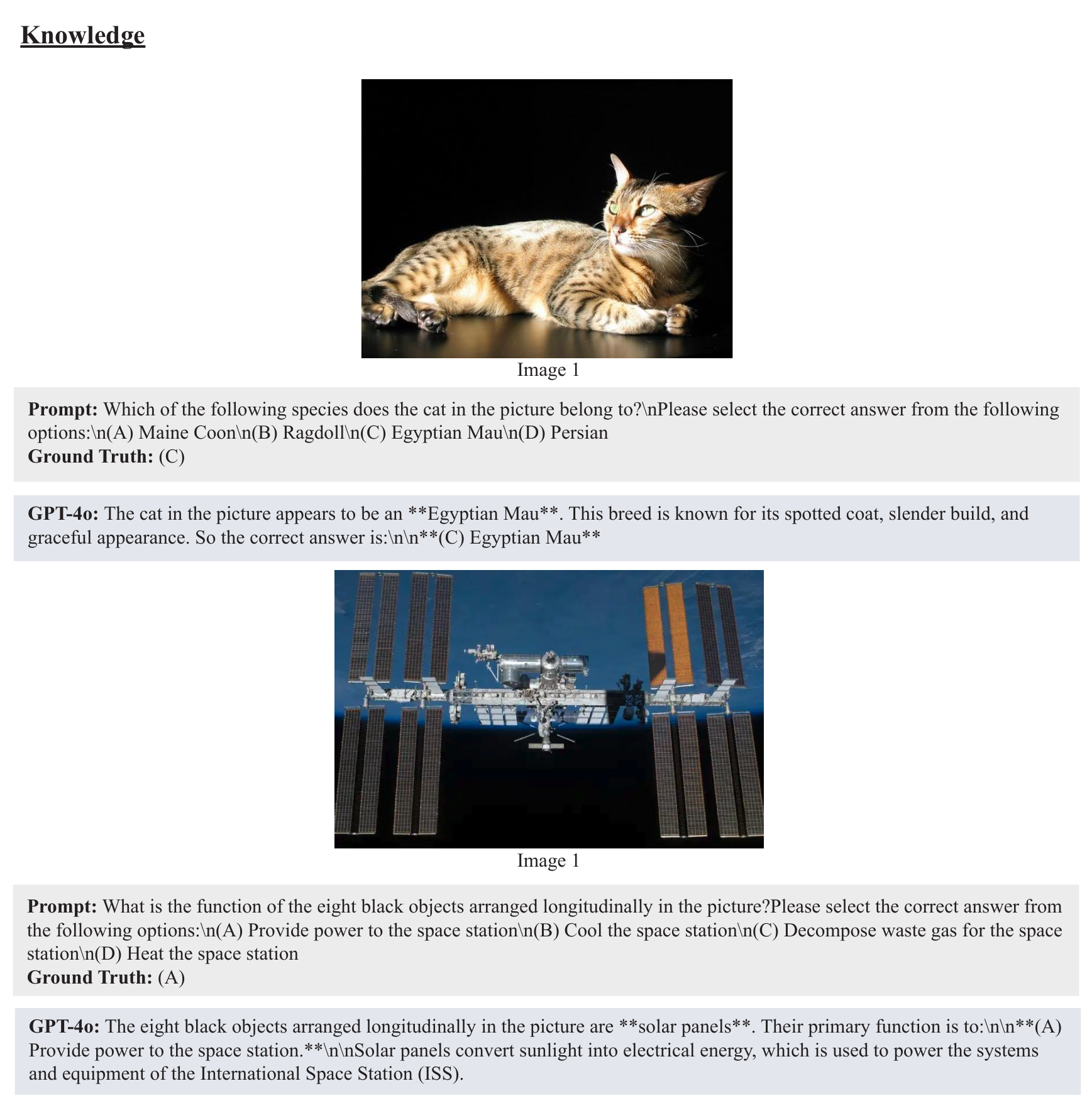}
    \caption{Examples of data in MANBench with GPT-4o predictions.(4/9)}
\end{figure*}
\newpage
\begin{figure*}[h]
    \centering
    \includegraphics[width=\textwidth]{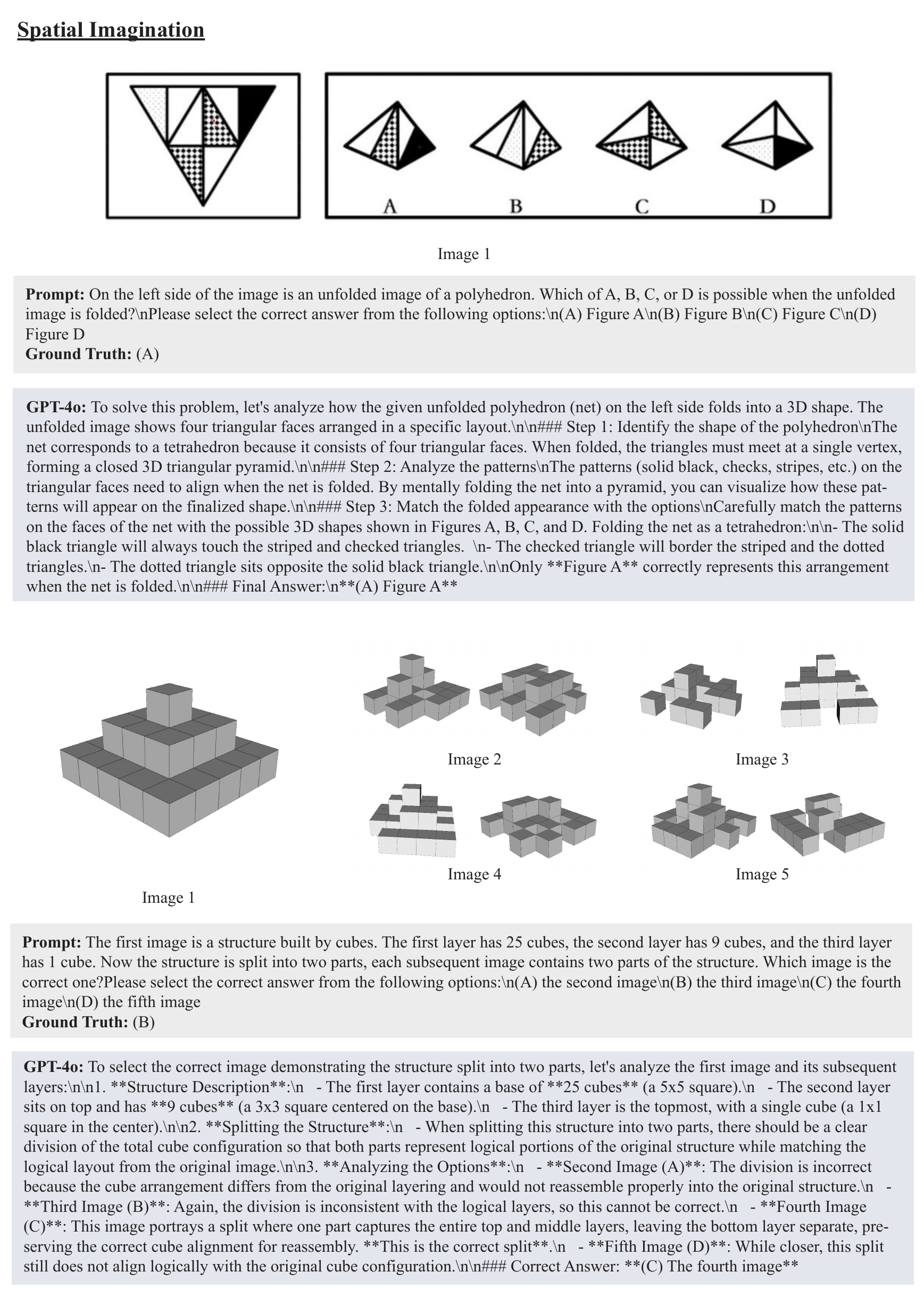}
    \caption{Examples of data in MANBench with GPT-4o predictions.(5/9)}
\end{figure*}
\newpage
\begin{figure*}[h]
    \centering
    \includegraphics[width=\textwidth]{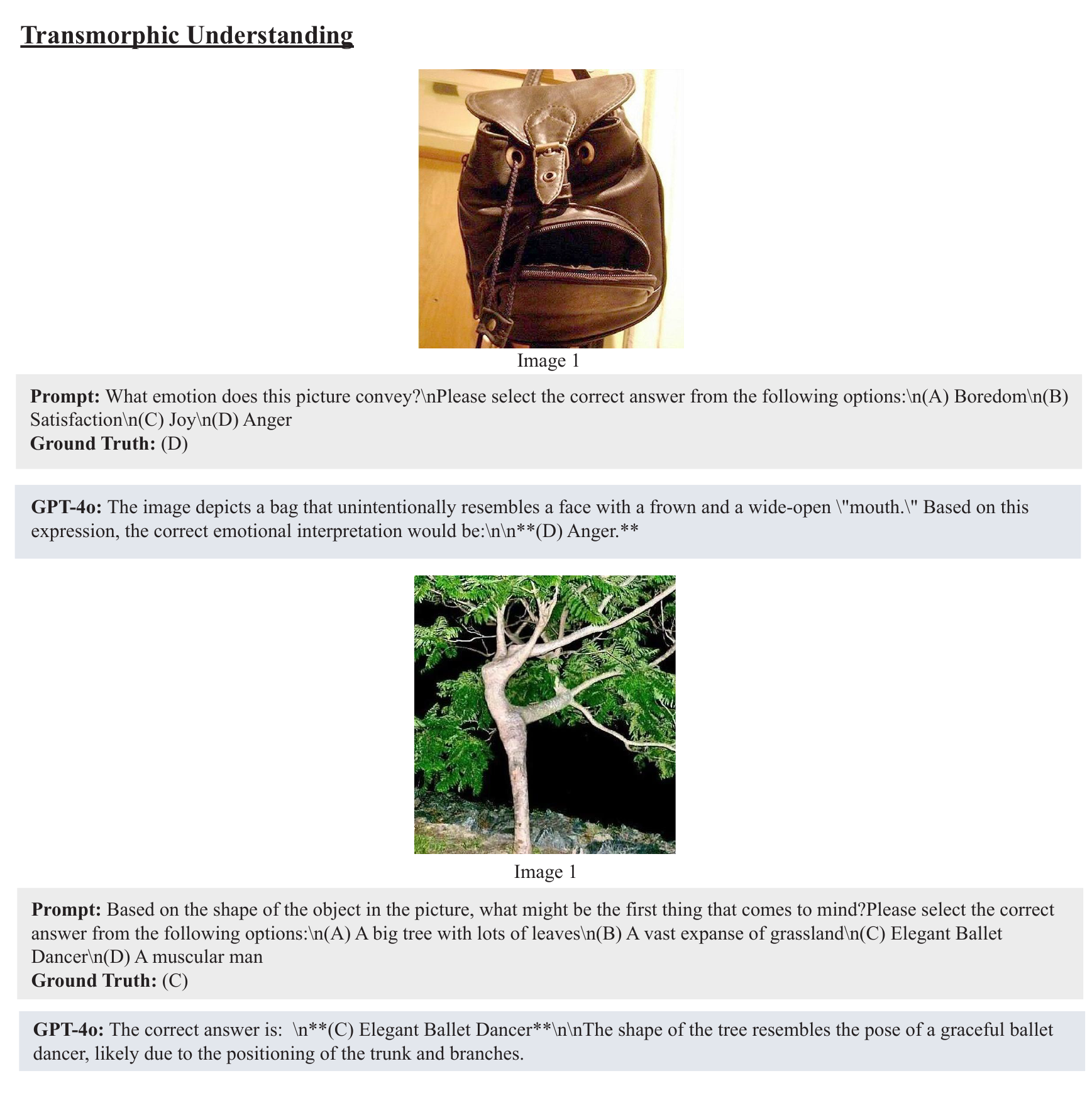}
    \caption{Examples of data in MANBench with GPT-4o predictions.(6/9)}
\end{figure*}
\newpage
\begin{figure*}[h]
    \centering
    \includegraphics[width=\textwidth]{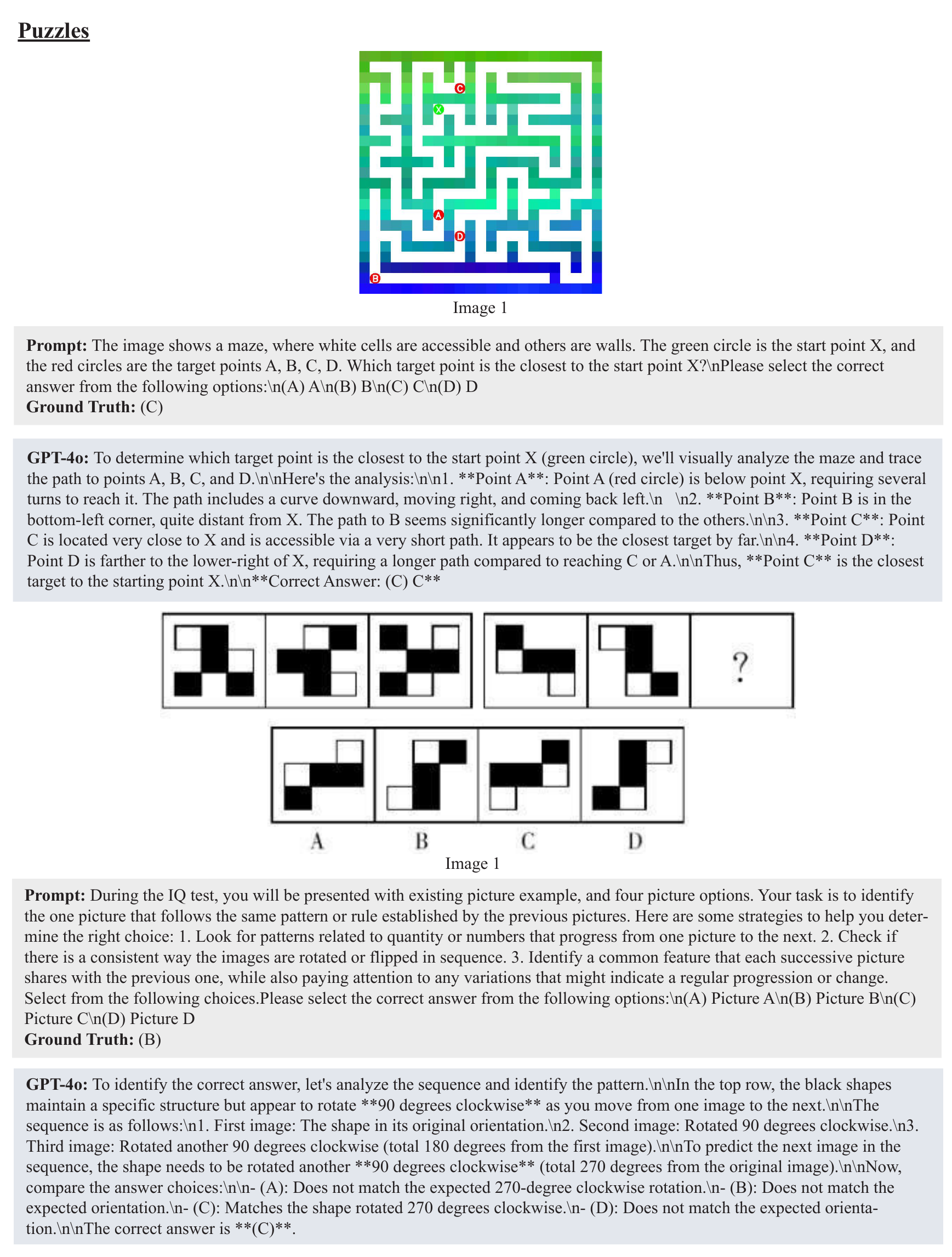}
    \caption{Examples of data in MANBench with GPT-4o predictions.(7/9)}
\end{figure*}
\newpage
\begin{figure*}[h]
    \centering
    \includegraphics[width=\textwidth]{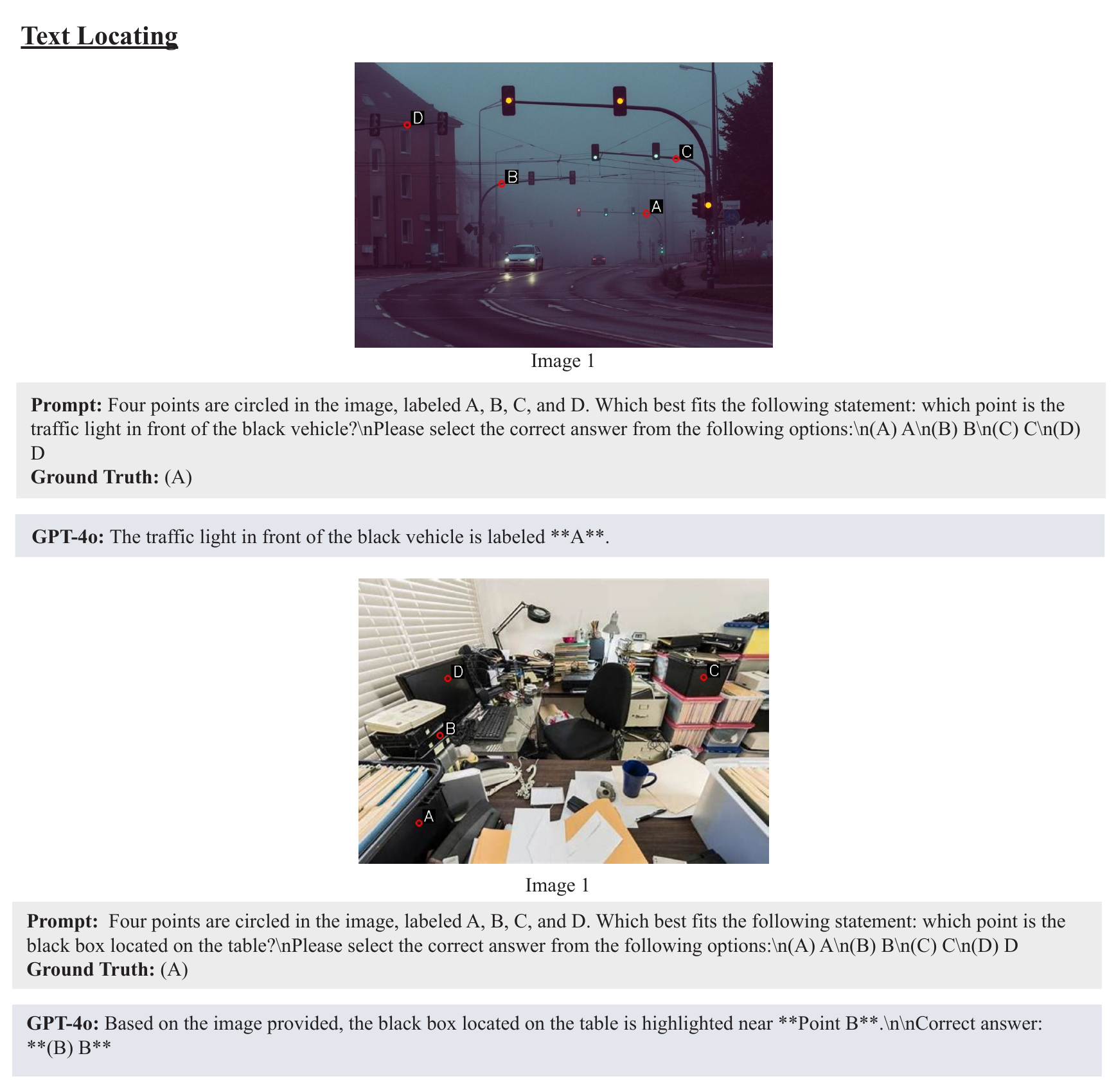}
    \caption{Examples of data in MANBench with GPT-4o predictions.(8/9)}
\end{figure*}
\newpage
\begin{figure*}[h]
    \centering
    \includegraphics[width=\textwidth]{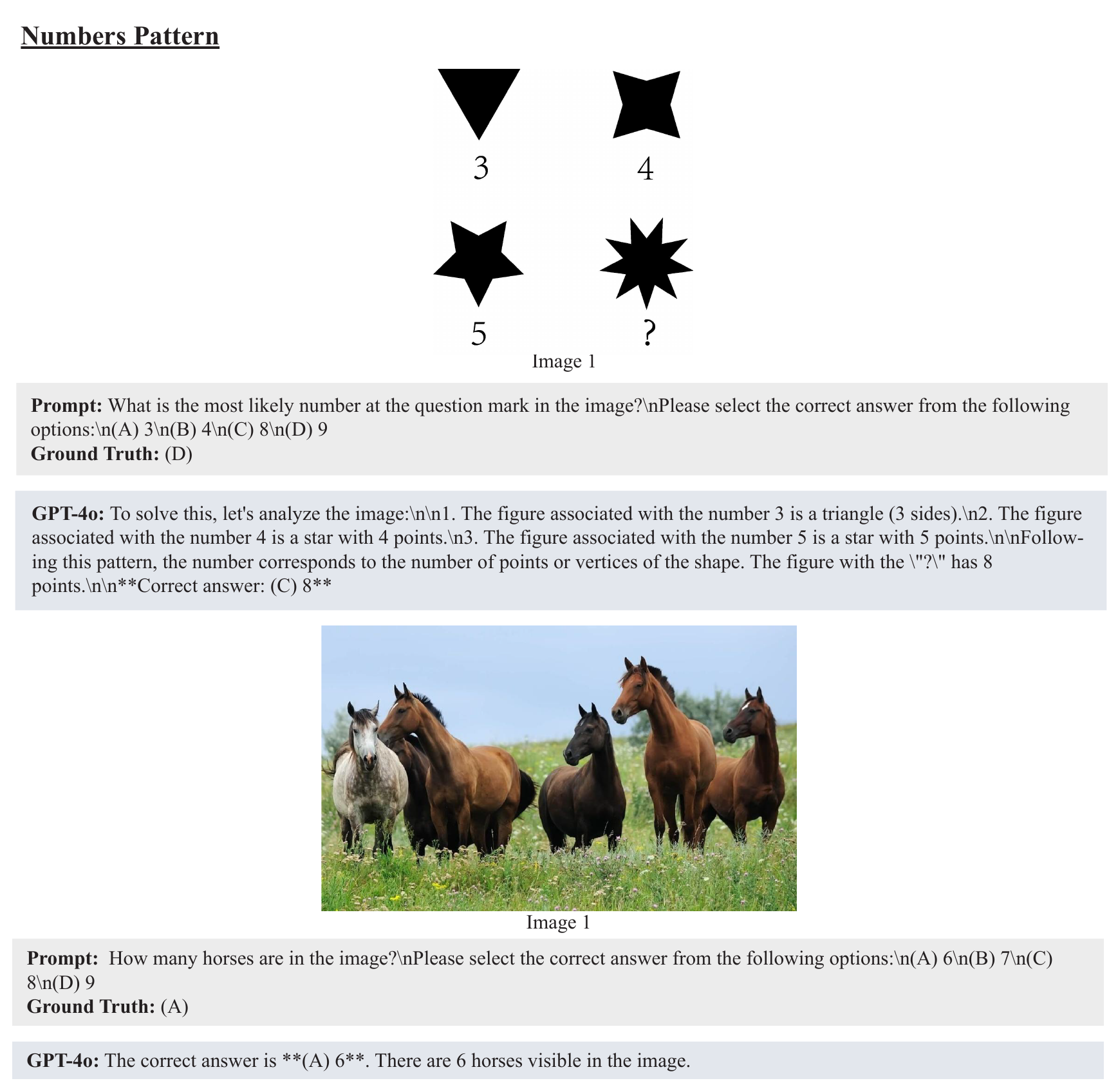}
    \caption{Examples of data in MANBench with GPT-4o predictions.(9/9)}
\end{figure*}
\end{document}